\documentclass{article}

\usepackage[preprint,nonatbib]{neurips_2020}
\usepackage[numbers]{natbib}

\usepackage[utf8]{inputenc} 
\usepackage[T1]{fontenc}    
\usepackage{hyperref}       
\usepackage{url}            
\usepackage{booktabs}       
\usepackage{amsfonts}       
\usepackage{nicefrac}       
\usepackage{microtype}      

\usepackage{graphicx}
\usepackage{xcolor}
\usepackage{subcaption}

\usepackage{amsmath,amssymb,amsthm}
\usepackage{physics}
\usepackage{mathtools}

\title{Conformal retrofitting via Riemannian manifolds: distilling task-specific graphs into pretrained embeddings}

\author{%
  Justin Dieter \\
  Stanford University \\
  \texttt{jdieter@cs.stanford.edu} \\
  \And
  Arun Tejasvi Chaganty \\
  Square Inc.\\
  \texttt{arun@squareup.com} \\
}

\begin{document}

\providecommand\sa{\ensuremath{\mathcal{a}}}
\providecommand\sd{\ensuremath{\mathcal{d}}}
\providecommand\se{\ensuremath{\mathcal{e}}}
\providecommand\sg{\ensuremath{\mathcal{g}}}
\providecommand\sh{\ensuremath{\mathcal{h}}}
\providecommand\si{\ensuremath{\mathcal{i}}}
\providecommand\sj{\ensuremath{\mathcal{j}}}
\providecommand\sk{\ensuremath{\mathcal{k}}}
\providecommand\sm{\ensuremath{\mathcal{m}}}
\providecommand\sn{\ensuremath{\mathcal{n}}}
\providecommand\so{\ensuremath{\mathcal{o}}}
\providecommand\sq{\ensuremath{\mathcal{q}}}
\providecommand\sr{\ensuremath{\mathcal{r}}}
\providecommand\st{\ensuremath{\mathcal{t}}}
\providecommand\su{\ensuremath{\mathcal{u}}}
\providecommand\sv{\ensuremath{\mathcal{v}}}
\providecommand\sw{\ensuremath{\mathcal{w}}}
\providecommand\sx{\ensuremath{\mathcal{x}}}
\providecommand\sy{\ensuremath{\mathcal{y}}}
\providecommand\sz{\ensuremath{\mathcal{z}}}
\providecommand\sA{\ensuremath{\mathcal{A}}}
\providecommand\sB{\ensuremath{\mathcal{B}}}
\providecommand\sC{\ensuremath{\mathcal{C}}}
\providecommand\sD{\ensuremath{\mathcal{D}}}
\providecommand\sE{\ensuremath{\mathcal{E}}}
\providecommand\sF{\ensuremath{\mathcal{F}}}
\providecommand\sG{\ensuremath{\mathcal{G}}}
\providecommand\sH{\ensuremath{\mathcal{H}}}
\providecommand\sI{\ensuremath{\mathcal{I}}}
\providecommand\sJ{\ensuremath{\mathcal{J}}}
\providecommand\sK{\ensuremath{\mathcal{K}}}
\providecommand\sL{\ensuremath{\mathcal{L}}}
\providecommand\sM{\ensuremath{\mathcal{M}}}
\providecommand\sN{\ensuremath{\mathcal{N}}}
\providecommand\sO{\ensuremath{\mathcal{O}}}
\providecommand\sP{\ensuremath{\mathcal{P}}}
\providecommand\sQ{\ensuremath{\mathcal{Q}}}
\providecommand\sR{\ensuremath{\mathcal{R}}}
\providecommand\sS{\ensuremath{\mathcal{S}}}
\providecommand\sT{\ensuremath{\mathcal{T}}}
\providecommand\sU{\ensuremath{\mathcal{U}}}
\providecommand\sV{\ensuremath{\mathcal{V}}}
\providecommand\sW{\ensuremath{\mathcal{W}}}
\providecommand\sX{\ensuremath{\mathcal{X}}}
\providecommand\sY{\ensuremath{\mathcal{Y}}}
\providecommand\sZ{\ensuremath{\mathcal{Z}}}
\providecommand\ba{\ensuremath{\mathbf{a}}}
\providecommand\bb{\ensuremath{\mathbf{b}}}
\providecommand\bc{\ensuremath{\mathbf{c}}}
\providecommand\bd{\ensuremath{\mathbf{d}}}
\providecommand\be{\ensuremath{\mathbf{e}}}
\providecommand\bg{\ensuremath{\mathbf{g}}}
\providecommand\bh{\ensuremath{\mathbf{h}}}
\providecommand\bi{\ensuremath{\mathbf{i}}}
\providecommand\bj{\ensuremath{\mathbf{j}}}
\providecommand\bk{\ensuremath{\mathbf{k}}}
\providecommand\bl{\ensuremath{\mathbf{l}}}
\providecommand\bn{\ensuremath{\mathbf{n}}}
\providecommand\bo{\ensuremath{\mathbf{o}}}
\providecommand\bp{\ensuremath{\mathbf{p}}}
\providecommand\bq{\ensuremath{\mathbf{q}}}
\providecommand\br{\ensuremath{\mathbf{r}}}
\providecommand\bs{\ensuremath{\mathbf{s}}}
\providecommand\bt{\ensuremath{\mathbf{t}}}
\providecommand\bu{\ensuremath{\mathbf{u}}}
\providecommand\bv{\ensuremath{\mathbf{v}}}
\providecommand\bw{\ensuremath{\mathbf{w}}}
\providecommand\bx{\ensuremath{\mathbf{x}}}
\providecommand\by{\ensuremath{\mathbf{y}}}
\providecommand\bz{\ensuremath{\mathbf{z}}}
\providecommand\bA{\ensuremath{\mathbf{A}}}
\providecommand\bB{\ensuremath{\mathbf{B}}}
\providecommand\bC{\ensuremath{\mathbf{C}}}
\providecommand\bD{\ensuremath{\mathbf{D}}}
\providecommand\bE{\ensuremath{\mathbf{E}}}
\providecommand\bF{\ensuremath{\mathbf{F}}}
\providecommand\bG{\ensuremath{\mathbf{G}}}
\providecommand\bH{\ensuremath{\mathbf{H}}}
\providecommand\bI{\ensuremath{\mathbf{I}}}
\providecommand\bJ{\ensuremath{\mathbf{J}}}
\providecommand\bK{\ensuremath{\mathbf{K}}}
\providecommand\bL{\ensuremath{\mathbf{L}}}
\providecommand\bM{\ensuremath{\mathbf{M}}}
\providecommand\bN{\ensuremath{\mathbf{N}}}
\providecommand\bO{\ensuremath{\mathbf{O}}}
\providecommand\bP{\ensuremath{\mathbf{P}}}
\providecommand\bQ{\ensuremath{\mathbf{Q}}}
\providecommand\bR{\ensuremath{\mathbf{R}}}
\providecommand\bS{\ensuremath{\mathbf{S}}}
\providecommand\bT{\ensuremath{\mathbf{T}}}
\providecommand\bU{\ensuremath{\mathbf{U}}}
\providecommand\bV{\ensuremath{\mathbf{V}}}
\providecommand\bW{\ensuremath{\mathbf{W}}}
\providecommand\bX{\ensuremath{\mathbf{X}}}
\providecommand\bY{\ensuremath{\mathbf{Y}}}
\providecommand\bZ{\ensuremath{\mathbf{Z}}}
\providecommand\Ba{\ensuremath{\mathbb{a}}}
\providecommand\Bb{\ensuremath{\mathbb{b}}}
\providecommand\Bc{\ensuremath{\mathbb{c}}}
\providecommand\Bd{\ensuremath{\mathbb{d}}}
\providecommand\Be{\ensuremath{\mathbb{e}}}
\providecommand\Bf{\ensuremath{\mathbb{f}}}
\providecommand\Bg{\ensuremath{\mathbb{g}}}
\providecommand\Bh{\ensuremath{\mathbb{h}}}
\providecommand\Bi{\ensuremath{\mathbb{i}}}
\providecommand\Bj{\ensuremath{\mathbb{j}}}
\providecommand\Bk{\ensuremath{\mathbb{k}}}
\providecommand\Bl{\ensuremath{\mathbb{l}}}
\providecommand\Bm{\ensuremath{\mathbb{m}}}
\providecommand\Bn{\ensuremath{\mathbb{n}}}
\providecommand\Bo{\ensuremath{\mathbb{o}}}
\providecommand\Bp{\ensuremath{\mathbb{p}}}
\providecommand\Bq{\ensuremath{\mathbb{q}}}
\providecommand\Br{\ensuremath{\mathbb{r}}}
\providecommand\Bs{\ensuremath{\mathbb{s}}}
\providecommand\Bt{\ensuremath{\mathbb{t}}}
\providecommand\Bu{\ensuremath{\mathbb{u}}}
\providecommand\Bv{\ensuremath{\mathbb{v}}}
\providecommand\Bw{\ensuremath{\mathbb{w}}}
\providecommand\Bx{\ensuremath{\mathbb{x}}}
\providecommand\By{\ensuremath{\mathbb{y}}}
\providecommand\Bz{\ensuremath{\mathbb{z}}}
\providecommand\BA{\ensuremath{\mathbb{A}}}
\providecommand\BB{\ensuremath{\mathbb{B}}}
\providecommand\BC{\ensuremath{\mathbb{C}}}
\providecommand\BD{\ensuremath{\mathbb{D}}}
\providecommand\BE{\ensuremath{\mathbb{E}}}
\providecommand\BF{\ensuremath{\mathbb{F}}}
\providecommand\BG{\ensuremath{\mathbb{G}}}
\providecommand\BH{\ensuremath{\mathbb{H}}}
\providecommand\BI{\ensuremath{\mathbb{I}}}
\providecommand\BJ{\ensuremath{\mathbb{J}}}
\providecommand\BK{\ensuremath{\mathbb{K}}}
\providecommand\BL{\ensuremath{\mathbb{L}}}
\providecommand\BM{\ensuremath{\mathbb{M}}}
\providecommand\BN{\ensuremath{\mathbb{N}}}
\providecommand\BO{\ensuremath{\mathbb{O}}}
\providecommand\BP{\ensuremath{\mathbb{P}}}
\providecommand\BQ{\ensuremath{\mathbb{Q}}}
\providecommand\BR{\ensuremath{\mathbb{R}}}
\providecommand\BS{\ensuremath{\mathbb{S}}}
\providecommand\BT{\ensuremath{\mathbb{T}}}
\providecommand\BU{\ensuremath{\mathbb{U}}}
\providecommand\BV{\ensuremath{\mathbb{V}}}
\providecommand\BW{\ensuremath{\mathbb{W}}}
\providecommand\BX{\ensuremath{\mathbb{X}}}
\providecommand\BY{\ensuremath{\mathbb{Y}}}
\providecommand\BZ{\ensuremath{\mathbb{Z}}}
\providecommand\balpha{\ensuremath{\mbox{\boldmath$\alpha$}}}
\providecommand\bbeta{\ensuremath{\mbox{\boldmath$\beta$}}}
\providecommand\btheta{\ensuremath{\mbox{\boldmath$\theta$}}}
\providecommand\bphi{\ensuremath{\mbox{\boldmath$\phi$}}}
\providecommand\bpi{\ensuremath{\mbox{\boldmath$\pi$}}}
\providecommand\bpsi{\ensuremath{\mbox{\boldmath$\psi$}}}
\providecommand\bmu{\ensuremath{\mbox{\boldmath$\mu$}}}
\providecommand\fig[1]{\begin{center} \includegraphics{#1} \end{center}}
\providecommand\Fig[4]{\begin{figure}[ht] \begin{center} \includegraphics[scale=#2]{#1} \end{center} \caption{\label{fig:#3} #4} \end{figure}}
\providecommand\FigTop[4]{\begin{figure}[t] \begin{center} \includegraphics[scale=#2]{#1} \end{center} \caption{\label{fig:#3} #4} \end{figure}}
\providecommand\FigStar[4]{\begin{figure*}[ht] \begin{center} \includegraphics[scale=#2]{#1} \end{center} \caption{\label{fig:#3} #4} \end{figure*}}
\providecommand\FigRight[4]{\begin{wrapfigure}{r}{0.5\textwidth} \begin{center} \includegraphics[width=0.5\textwidth]{#1} \end{center} \caption{\label{fig:#3} #4} \end{wrapfigure}}
\providecommand\aside[1]{\quad\text{[#1]}}
\providecommand\interpret[1]{\llbracket #1 \rrbracket} 
\providecommand{\var}{\text{Var}} 
\providecommand{\cov}{\text{Cov}} 
\providecommand\p[1]{\ensuremath{\left( #1 \right)}} 
\providecommand\pa[1]{\ensuremath{\left\langle #1 \right\rangle}} 
\providecommand\pb[1]{\ensuremath{\left[ #1 \right]}} 
\providecommand\pc[1]{\ensuremath{\left\{ #1 \right\}}} 
\providecommand\eval[2]{\ensuremath{\left. #1 \right|_{#2}}} 
\providecommand\inv[1]{\ensuremath{\frac{1}{#1}}}
\providecommand\half{\ensuremath{\frac{1}{2}}}
\providecommand\R{\ensuremath{\mathbb{R}}} 
\providecommand\Z{\ensuremath{\mathbb{Z}}} 
\providecommand\inner[2]{\ensuremath{\left< #1, #2 \right>}} 
\providecommand\mat[2]{\ensuremath{\left(\begin{array}{#1}#2\end{array}\right)}} 
\providecommand\eqn[1]{\begin{align} #1 \end{align}} 
\providecommand\eqnl[2]{\begin{align} \label{eqn:#1} #2 \end{align}} 
\providecommand\eqdef{\ensuremath{\stackrel{\rm def}{=}}} 
\providecommand{\1}{\mathbb{I}} 
\providecommand{\bone}{\mathbf{1}} 
\providecommand{\bzero}{\mathbf{0}} 
\providecommand\refeqn[1]{(\ref{eqn:#1})}
\providecommand\refeqns[2]{(\ref{eqn:#1}) and (\ref{eqn:#2})}
\providecommand\refchp[1]{Chapter~\ref{chp:#1}}
\providecommand\refchap[1]{Chapter~\ref{chap:#1}}
\providecommand\refsec[1]{Section~\ref{sec:#1}}
\providecommand\refsecs[2]{Sections~\ref{sec:#1} and~\ref{sec:#2}}
\providecommand\reffig[1]{Figure~\ref{fig:#1}}
\providecommand\reffigs[2]{Figures~\ref{fig:#1} and~\ref{fig:#2}}
\providecommand\reffigss[3]{Figures~\ref{fig:#1},~\ref{fig:#2}, and~\ref{fig:#3}}
\providecommand\reffigsss[4]{Figures~\ref{fig:#1},~\ref{fig:#2},~\ref{fig:#3}, and~\ref{fig:#4}}
\providecommand\reftab[1]{Table~\ref{tab:#1}}
\providecommand\refapp[1]{Appendix~\ref{sec:#1}}
\providecommand\refthm[1]{Theorem~\ref{thm:#1}}
\providecommand\refthms[2]{Theorems~\ref{thm:#1} and~\ref{thm:#2}}
\providecommand\reflem[1]{Lemma~\ref{lem:#1}}
\providecommand\reflems[2]{Lemmas~\ref{lem:#1} and~\ref{lem:#2}}
\providecommand\refalg[1]{Algorithm~\ref{alg:#1}}
\providecommand\refalgs[2]{Algorithms~\ref{alg:#1} and~\ref{alg:#2}}
\providecommand\refex[1]{Example~\ref{ex:#1}}
\providecommand\refexs[2]{Examples~\ref{ex:#1} and~\ref{ex:#2}}
\providecommand\refprop[1]{Proposition~\ref{prop:#1}}
\providecommand\refdef[1]{Definition~\ref{def:#1}}
\providecommand\refcor[1]{Corollary~\ref{cor:#1}}
\providecommand\Chapter[2]{\chapter{#2}\label{chp:#1}}
\providecommand\Section[2]{\section{#2}\label{sec:#1}}
\providecommand\Subsection[2]{\subsection{#2}\label{sec:#1}}
\providecommand\Subsubsection[2]{\subsubsection{#2}\label{sec:#1}}
\ifthenelse{\isundefined{\definition}}{\newtheorem{definition}{Definition}}{}
\ifthenelse{\isundefined{\assumption}}{\newtheorem{assumption}{Assumption}}{}
\ifthenelse{\isundefined{\hypothesis}}{\newtheorem{hypothesis}{Hypothesis}}{}
\ifthenelse{\isundefined{\proposition}}{\newtheorem{proposition}{Proposition}}{}
\ifthenelse{\isundefined{\theorem}}{\newtheorem{theorem}{Theorem}}{}
\ifthenelse{\isundefined{\lemma}}{\newtheorem{lemma}{Lemma}}{}
\ifthenelse{\isundefined{\corollary}}{\newtheorem{corollary}{Corollary}}{}
\ifthenelse{\isundefined{\alg}}{\newtheorem{alg}{Algorithm}}{}
\ifthenelse{\isundefined{\example}}{\newtheorem{example}{Example}}{}
\providecommand\cv{\ensuremath{\to}} 
\providecommand\cvL{\ensuremath{\xrightarrow{\mathcal{L}}}} 
\providecommand\cvd{\ensuremath{\xrightarrow{d}}} 
\providecommand\cvP{\ensuremath{\xrightarrow{P}}} 
\providecommand\cvas{\ensuremath{\xrightarrow{a.s.}}} 
\providecommand\eqdistrib{\ensuremath{\stackrel{d}{=}}} 
\providecommand{\E}{\ensuremath{\mathbb{E}}} 
\providecommand\KL[2]{\ensuremath{\text{KL}\left( #1 \| #2 \right)}} 

\providecommand{\NN}{\mathbb{N}}
\providecommand{\ZZ}{\mathbb{Z}}
\providecommand{\QQ}{\mathbb{Q}}
\providecommand{\II}{\mathbf{I}}
\providecommand{\R}{\mathbb{R}}
\providecommand{\X}{\mathcal{X}}
\providecommand{\Y}{\mathcal{Y}}
\providecommand{\LL}{\mathcal{L}}
\providecommand{\E}{\mathbb{E}}
\providecommand{\C}{\mathbb{C}}
\providecommand{\0}{\mathbf{0}}
\providecommand{\eps}{\varepsilon}
\providecommand{\TT}{\mathscr{T}}
\providecommand{\BB}{\mathscr{B}}
\providecommand{\CCC}{\mathscr{C}}
\providecommand{\UU}{\mathscr{U}}
\providecommand{\FF}{\mathscr{F}}
\providecommand{\HH}{\mathbb{H}}
\providecommand{\N}{\mathcal{N}}
\providecommand{\Z}{\mathbb{Z}}
\providecommand{\OK}{\mathcal{O}_K}
\providecommand{\OL}{\mathcal{O}_L}
\providecommand{\bfm}[1]{\mathbf{#1}}

\providecommand{\ac}[1]{{\color{red}[AC:\@ #1]}}
\providecommand{\jd}[1]{{\color{blue}[JD:\@ #1]}}
\providecommand{\todo}[1]{{\color{red}{[#1]}}}
\providecommand{\needscite}{{\color{yellow}{$^\text{[citation needed]}$}}}

\providecommand{\WordNet}{{\textsc{WordNet}}}
\providecommand{\ConceptNet}{{\textsc{ConceptNet}}}
\providecommand{\RIsoN}{{\textsc{RIsoN}}}

\renewcommand{\inv}{\ensuremath{^{-1}}}
\providecommand{\pinv}{\ensuremath{^{\dagger}}}
\renewcommand{\Re}{\mathbb{R}}
\providecommand{\mEuc}{\ensuremath{\mathbb{E}}}
\providecommand{\mSph}{\ensuremath{\mathbb{S}}}
\providecommand{\mHyp}{\ensuremath{\mathbb{H}}}
\providecommand{\mS}{\sS}
\providecommand{\mT}{\sT}
\providecommand{\inS}{\ensuremath{^{\mS}}}
\providecommand{\inT}{\ensuremath{^{\mT}}}
\providecommand{\pullback}[1]{{\bar{#1}}}
\providecommand{\pushforward}{{\partial}}
\providecommand{\retr}{\operatorname{retr}}

\providecommand{\loss}{\ell}
\providecommand{\fidelity}{\text{fid}}
\providecommand{\preservation}{\text{pre}}
\providecommand{\conformal}{\text{conf}}
\providecommand{\isometry}{\text{iso}}
\providecommand{\prox}{\text{prox}}

\providecommand{\DMammals}{\textsc{Mammals}}
\providecommand{\DNouns}{\textsc{Nouns}}

\providecommand{\sretro}{\textsc{SR}}
\providecommand{\eretro}{\textsc{ER}}
\providecommand{\cretro}{\textsc{CR}}

\providecommand{\Neighborhood}{\operatorname{N}}
\providecommand{\Margin}{\lambda}
\providecommand{\ReLU}{\operatorname{ReLU}}

\providecommand{\T}{\ensuremath{^\top}}

\providecommand{\avg}{{\text{avg}}}

\providecommand{\circled}[1]{{\raisebox{.5pt}{\tiny \textcircled{\raisebox{-.9pt} {#1}}}}}

\providecommand{\RFL}{\ensuremath{\operatorname{RFL}}}
\providecommand{\EFL}{\ensuremath{\operatorname{EFL}}}

\maketitle

\begin{abstract}
  Pretrained (language) embeddings are versatile, task-agnostic feature representations of entities, like words, that are central to many machine learning applications. 
These representations can be enriched through \textit{retrofitting}, a class of methods that incorporate task-specific domain knowledge encoded as a graph over a subset of these entities.
However, existing retrofitting algorithms face two limitations: they overfit the observed graph by failing to represent relationships with missing entities; and they underfit the observed graph by only learning embeddings in Euclidean manifolds, which cannot faithfully represent even simple tree-structured or cyclic graphs.
We address these problems with two key contributions:
(i) we propose a novel regularizer, a \textit{conformality regularizer}, that preserves local geometry from the pretrained embeddings---enabling generalization to missing entities and
(ii) a new Riemannian feedforward layer that learns to map pre-trained embeddings onto a non-Euclidean manifold that can better represent the entire graph.
Through experiments on WordNet, we demonstrate that the conformality regularizer prevents even existing (Euclidean-only) methods from overfitting on link prediction for missing entities, and---together with the Riemannian feedforward layer---learns non-Euclidean embeddings that outperform them.

\end{abstract}

\section{\label{sec:intro} Introduction}

\begin{figure*}[t]
  \begin{subfigure}[t]{0.24\textwidth}
    \includegraphics[width=\linewidth]{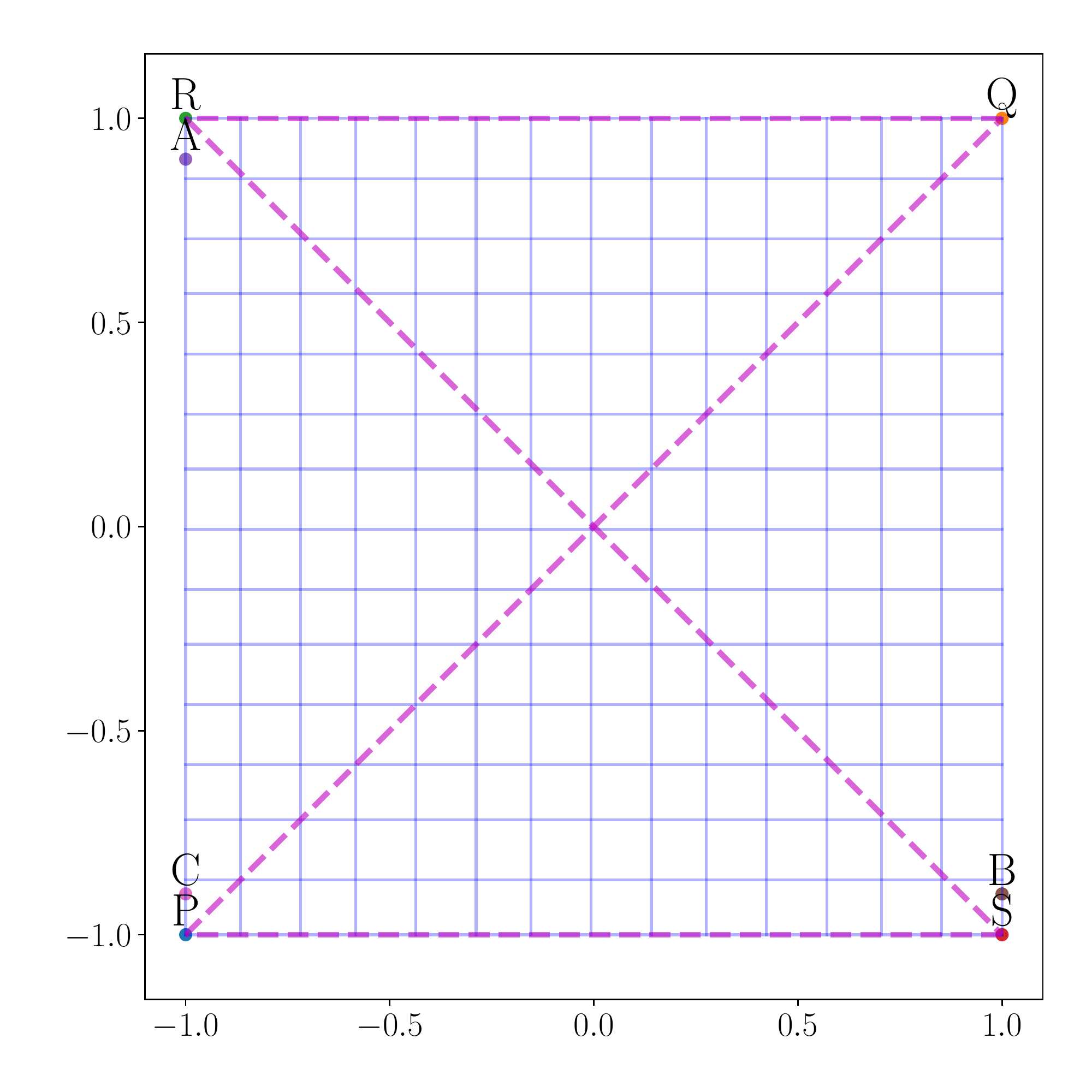}
    \caption{\label{fig:crossed-input} Embeddings and graph}
  \end{subfigure}
  \begin{subfigure}[t]{0.24\textwidth}
    \includegraphics[width=\linewidth]{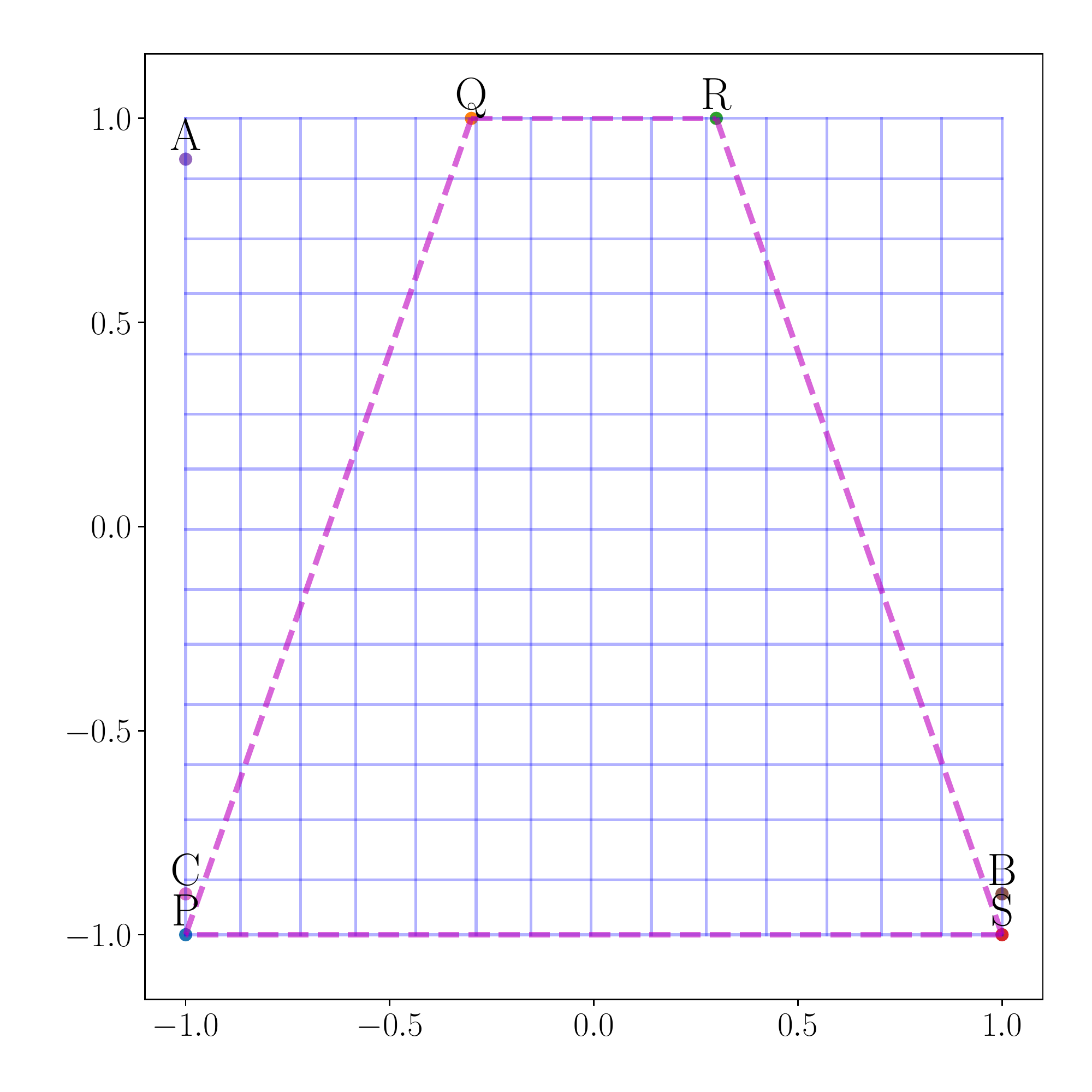}
    \caption{\label{fig:crossed-retrofitting} Standard Retrofitting}
  \end{subfigure}
  \begin{subfigure}[t]{0.24\textwidth}
    \includegraphics[width=\linewidth]{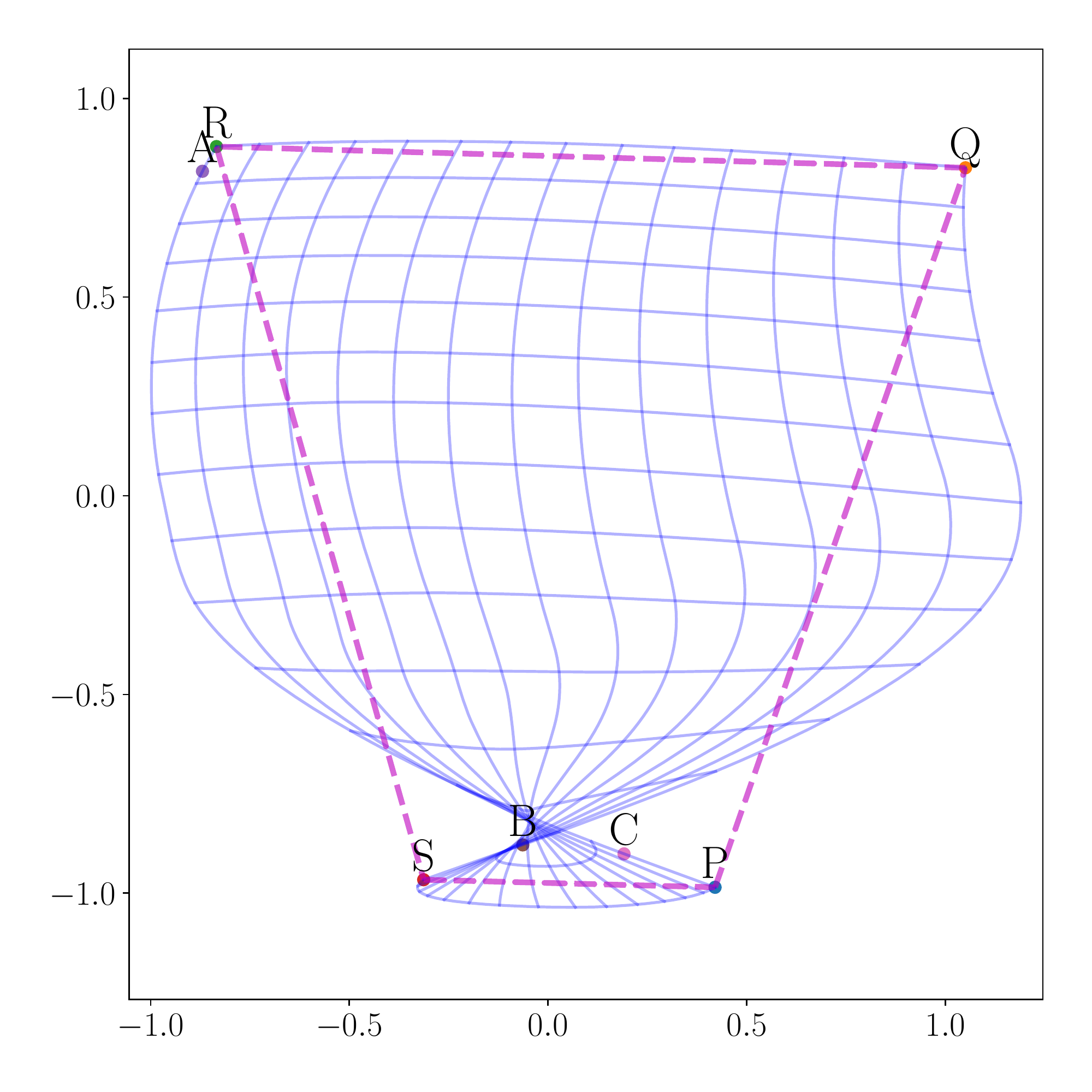}
    \caption{\label{fig:crossed-proximal} Explicit Retrofitting}
  \end{subfigure}
  \begin{subfigure}[t]{0.24\textwidth}
    \includegraphics[width=\linewidth]{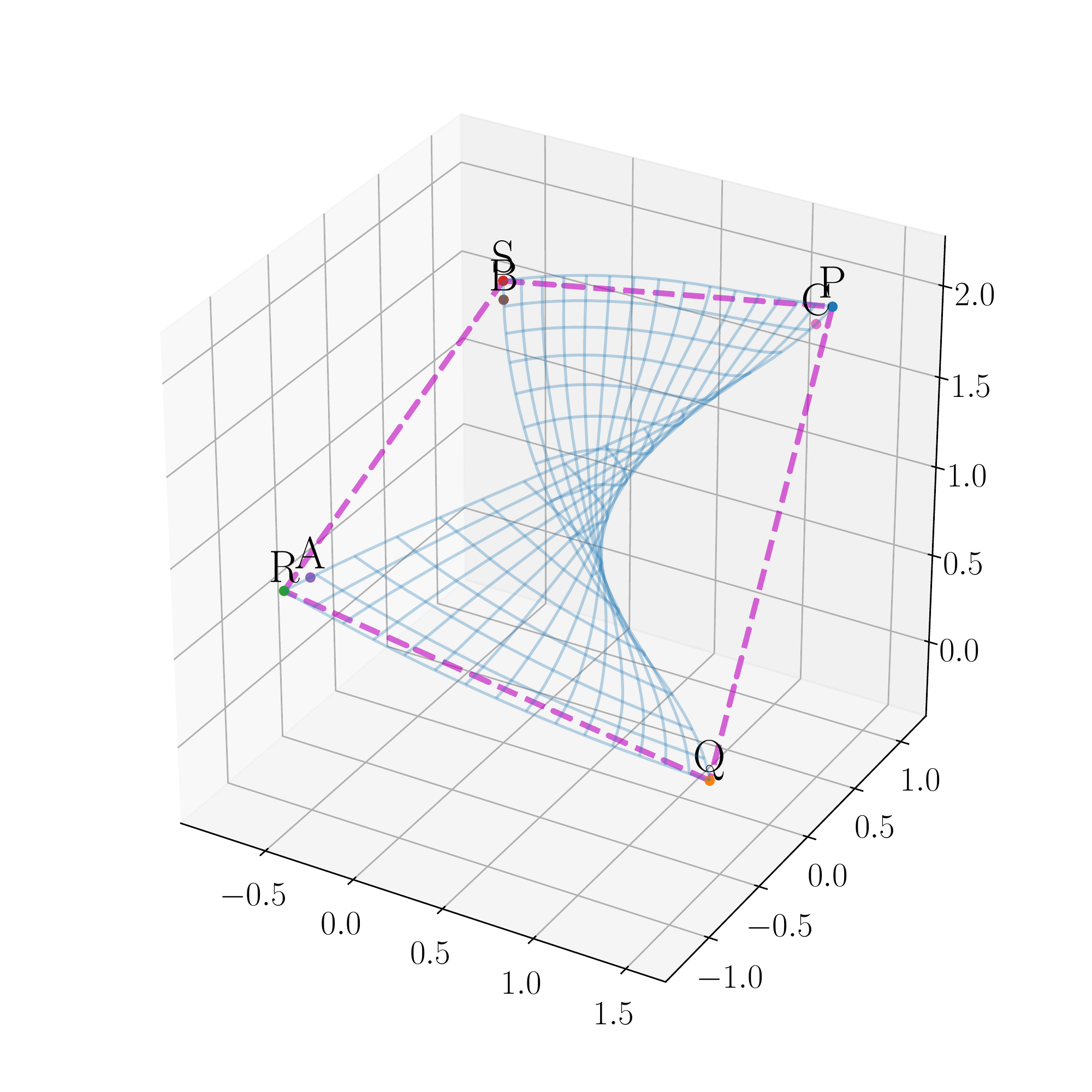}
    \caption{\label{fig:crossed-isometry} Conformal Retrofitting}
  \end{subfigure}
  \caption{\label{fig:crossed} 
  \textit{Comparison of embeddings learned by various retrofitting methods.}
  Retrofitting methods transform pretrained embeddings to better represent a task-specific graph.
  Panel (a) illustrates an example of two-dimensional pretrained embeddings (ABC and PQRS), overlaid with the task-specific graph over entities PQRS;\@ the entities ABC are missing from this graph.
  According to the graph, retrofitting methods should move P and Q closer to each other without also bringing P and R or Q and S together.
  (b) Standard retrofitting solves this problem by only moving the observed vertices, breaking the existing relationship between A and R, and failing to learn the one implied between A and B.
  (c) Explicit retrofitting learns a similar transformation using a continuous map that also applies to unobserved entities; now the relationship between A and R is preserved, but B and C have been forced together because the learned map does not preserve the geometry of the pretrained embeddings.
  (d) \textit{Conformal retrofitting} introduces a \textit{conformality} regularizer that explicitly preserves distances and angles; it results in a smoother transformation, resembling a saddle, which learns the relationship between A and B and preserves the (non-)relationship between B and C.
  }
\end{figure*}

Pretrained embeddings~\cite{mikolov2013distributed,pennington2014glove,donahue2014decaf,razavian2014cnn} and models~\cite{krizhevsky2012imagenet,he2016deep,szegedy2016rethinking,szegedy2017inception,radford2018improving,devlin2019bert,liu2019roberta} underpin many state-of-the-art results in computer vision and natural language processing across a variety of tasks and domains.
These embeddings and models can be further improved for specific domains by using task-specific information, often encoded as a graph:
  word embeddings and language models better represent semantic similarity (as opposed to just distributional similarity) when combined with lexical ontologies~\cite{mrksic2017semantic};
  image classifiers can generalize to new or rare classes when combined with knowledge graphs~\cite{wang2018zeroshot,peng2019fewshot,chen2019knowledge};
  medical diagnoses can be improved when combined with medical knowledge graphs~\cite{li2019improving};
  databases can impute missing data better when embeddings are specialized to their relational data~\cite{gunther2020retro}.

Retrofitting methods incorporate these task-specific graphs either by
  directly translating the embeddings (standard retrofitting)~\cite{faruqui2015retrofitting,mrksic2017semantic,rothe2017autoextend,glavas2019monolingual} or 
  by learning a neural network to do the same (explicit retrofitting)~\cite{glavas2018explicit,kamath2019specializing}.
Both standard and explicit retrofitting represent new relationships between entities observed in the task-specific graph;
  however, it is important to consider their impact on \textit{unobserved entities} because most task-specific graphs are characteristically incomplete (\reffig{crossed}).
Standard retrofitting methods do not translate embeddings for unobserved entities and hence cannot capture relationships for these entities (\reffig{crossed-retrofitting}).
Explicit retrofitting methods do not preserve the geometry of pretrained embeddings and hence unintentionally modify relationships to and between unobserved entities (\reffig{crossed-proximal}).
We regard this as a problem of overfitting.


In this paper, we address the overfitting problem with a novel regularizer, a \textit{conformality regularizer}, based on the pullback metric from Riemannian geometry.
The regularizer directly encourages the learned transformation to preserve distances and angles around each embedding point, resulting in smoother transformations (\reffig{crossed-isometry}).
We also introduce a single hyper-parameter that allows for bounded distortions in distances while still preserving angles; our best results were obtained when moderate distortions are allowed.

The conformality regularizer naturally extends to non-Euclidean Riemannian manifolds;
  these manifolds have been shown to better represent graphs than the Euclidean manifolds used by existing retrofitting methods~\cite{nickel2017poincare,gu2019learning,balazevic2019multi}.
In fact, it is well-known that Euclidean manifolds require a logarithmic number of dimensions to represent even simple tree-structured graphs~\cite{matouvsek1999embedding}, while hyperbolic manifolds can represent such graphs with just two dimensions.
We propose a new \textit{Riemannian feedforward layer} that extends the conventional feedforward layer to arbitrary heterogeneous input and output manifolds.
We use this new layer to transform pretrained embeddings from their Euclidean manifolds to a target Riemannian manifold.

We evaluate on predicting links to held out words from WordNet, a popular lexical knowledge graph used in a variety of language and vision tasks. 
On this task, we show that conformal regularization provides significant improvements to existing explicit retrofitting methods.
Moreover, when combined with the proposed Riemannian feedforward layer, we find that non-Euclidean product manifolds such as $\mSph^{30}\times\mHyp^{30}$ improve link prediction scores not only for both held-out entities, but for entities observed during training time too.

In summary, our contributions are:
\begin{enumerate}
  \item a novel regularization technique, conformality regularization, based on the pullback metric that better preserves angles and distances. 
  \item a new \textit{Riemannian feedforward layer} that can operate on heterogeneous input and output manifolds.
  \item experiments that show that conformal regularization improves the generalizability of existing explicit retrofitting methods and that product manifolds lead to better retrofitting.
\end{enumerate}

\section{\label{sec:background} Setup}

Let us now setup notation and formally define retrofitting as a task.
Our definitions expand prior work to apply when the pretrained and retrofitted embeddings lie on non-Euclidean Riemannian manifolds.
We begin by reviewing some key concepts from Riemannian geometry.

\subsection{\label{sec:background-manifolds} Riemannian manifolds}
A $d$-dimensional Riemannian manifold $(M, g)$ is a smooth manifold $M$ with an inner-product \textit{metric} $g$; 
  for any point $p \in M$, there exists a \text{tangent space} $T_p M$ that is isomorphic to $d$-dimensional Euclidean space $\Re^d$.
The metric $g_p: T_p M \times T_p M \to \Re$ smoothly varies with $p \in M$.
Key examples of Riemannian manifolds include
  the Euclidean manifold $\mEuc$ with $g^{\mEuc}_p(x, y) = x\T y$;
  the Poincare ball (an instance of a hyperbolic manifold) $\mHyp$ with $g^{\mHyp}_p(x, y) = \frac{2 x\T y}{1 - \|p\|^2}$; and
  the Spherical manifold with $g^{\mSph}_p(x, y) = \frac{x\T y}{\|x\|\|y\|}$.
Tangent spaces are regular vector spaces, allowing us to conveniently represent $g_p$ as a matrix, also known as the metric tensor:
$G^\mEuc_p = I$, $G^\mHyp_p = \frac{2}{1 - \|p\|^2} I$.

The (geodesic) distance between two points on the manifold, $d(x, y)$, is defined to be the shortest path length between the two points:
$d(x, y) \eqdef \min_\gamma \int_{0}^{1} \sqrt{g_{\gamma(t)}(\dot{\gamma}(t), \dot{\gamma}(t))} dt$,
where $\gamma: [0, 1] \to M$ is any smooth curve starting at $x$ and ending at $y$.
We focus on Euclidean (\mEuc), spherical (\mSph) and hyperbolic (\mHyp) manifolds and their products (defined below) which have efficient, closed-form solutions to the geodesic equation above.

Given two manifolds $(M, g)$ and $(N, g')$, their product manifold $M \times N$ has the following metric:
$\bar{g}_{(p, p')}\left((x, x'), (y, y') \right) = g_{p}(x, y) + g'_{p'}(x', y')$,
where $p \in M$, $x, y \in T_{p} M$, $p' \in N$ and $x', y' \in T_{p'} N$.
Consequently, the geodesic on the product manifold also decomposes:
$\bar{d}{((x, x'), (y, y'))}^2 = d{(x, y)}^2 + d'{(x', y')}^2$.

Finally, given $p \in M$, one can move to and from the tangent space using the $\log$ and $\exp$ maps:
  for $q \in M$, the logarithmic map $\log_p(q)$ returns a vector $v \in T_p M$ along the geodesic between the $p$ and $q$ of length $d(p, q)$;
  for $v \in T_p M$, the exponential map $\exp_p(v) \to q$ returns a point $q \in M$ obtained by following the geodesic in the direction of $v$ for a distance of $\|v\|$.
$\log_p$ and $\exp_q$ are inverses of each other only when $p = q$; 
  $\log_p(q)$ and $\exp_p(v)$ is differentiable in $p$, $q$ and $v$.

\subsection{\label{sec:background-retrofitting} Retrofitting}

Let $S$ be a pretrained source embedding in a given manifold $(\sS, g\inS)$ and $\sG(V, E)$ be the given (task-specific) graph,
  and $T$ be the retrofitted target embeddings in a chosen target manifold $(\sT, g\inT)$.
We define $V\inS = \{v\inS \vert v \in V\} \subset S$ to be the associated input embeddings of the vertices of the graph,
  and $V\inT = \{v\inT \vert v \in V\} \subset T$ their corresponding target embeddings.

The goal of retrofitting is learn target embeddings $T$ that:
  (a) faithfully represents distances in $\sG$ and 
  (b) preserves the geometry of the source embeddings $S$.
These two properties are typically captured by a multi-objective loss function using respective fidelity $\loss_\fidelity$ and preservation $\loss_\preservation$ loss terms:
\begin{align}
  \sL &= \sum_{u, v \in E} \loss_\fidelity(u, v) + \lambda \sum_{w\in V} \loss_\preservation(w), \label{eqn:retro}
\end{align}
where $\lambda$ is a hyperparameter that balances the two objectives.
Standard retrofitting~\cite{faruqui2015retrofitting} directly learns the target embeddings $V\inT$ with the following objective: 
\begin{align}
  \sL^\sretro(V\inT) &= 
    \sum_{(u, v) \in E} \underbrace{\|u\inT - v\inT\|^2}_{\loss_\fidelity^\sretro(u, v)} +
    \lambda \sum_{w \in V} \underbrace{\|w\inT - w\inS\|^2}_{\loss_\preservation^\sretro(w)}. \label{eqn:sretro}
\end{align}
Explicit retrofitting~\cite{glavas2018explicit} instead learns a neural network to transform the input embeddings, $v\inT = f_\theta(v\inS)$,
with graph-distance based fidelity loss:
\begin{align}
  \sL^{\eretro}(\theta) &= 
  \sum_{(u, v) \in E} \underbrace{\|d(u\inT, v\inT) - d_\sG(u, v)\|^2}_{\loss_\fidelity^\eretro(u, v)} +
    \lambda \sum_{w \in V} \underbrace{\|w\inT - w\inS\|^2}_{\loss_\preservation^\eretro(w)}, \label{eqn:eretro}
\end{align}
where $d(u\inT, v\inT)$ is one of Euclidean or cosine distance, and $d_\sG(u, v)$ is the graph distance between the two vertices.
Both methods share the same preservation loss, which we call \textit{proximity regularization}: $\loss_\prox(w) = \| w\inS - w\inT \|^2$.

\section{\label{sec:cretro} Conformal retrofitting}

\subsection{\label{sec:creg} Conformality regularization}
\begin{figure*}[t]
  \begin{subfigure}[t]{0.24\textwidth}
    \includegraphics[width=\linewidth]{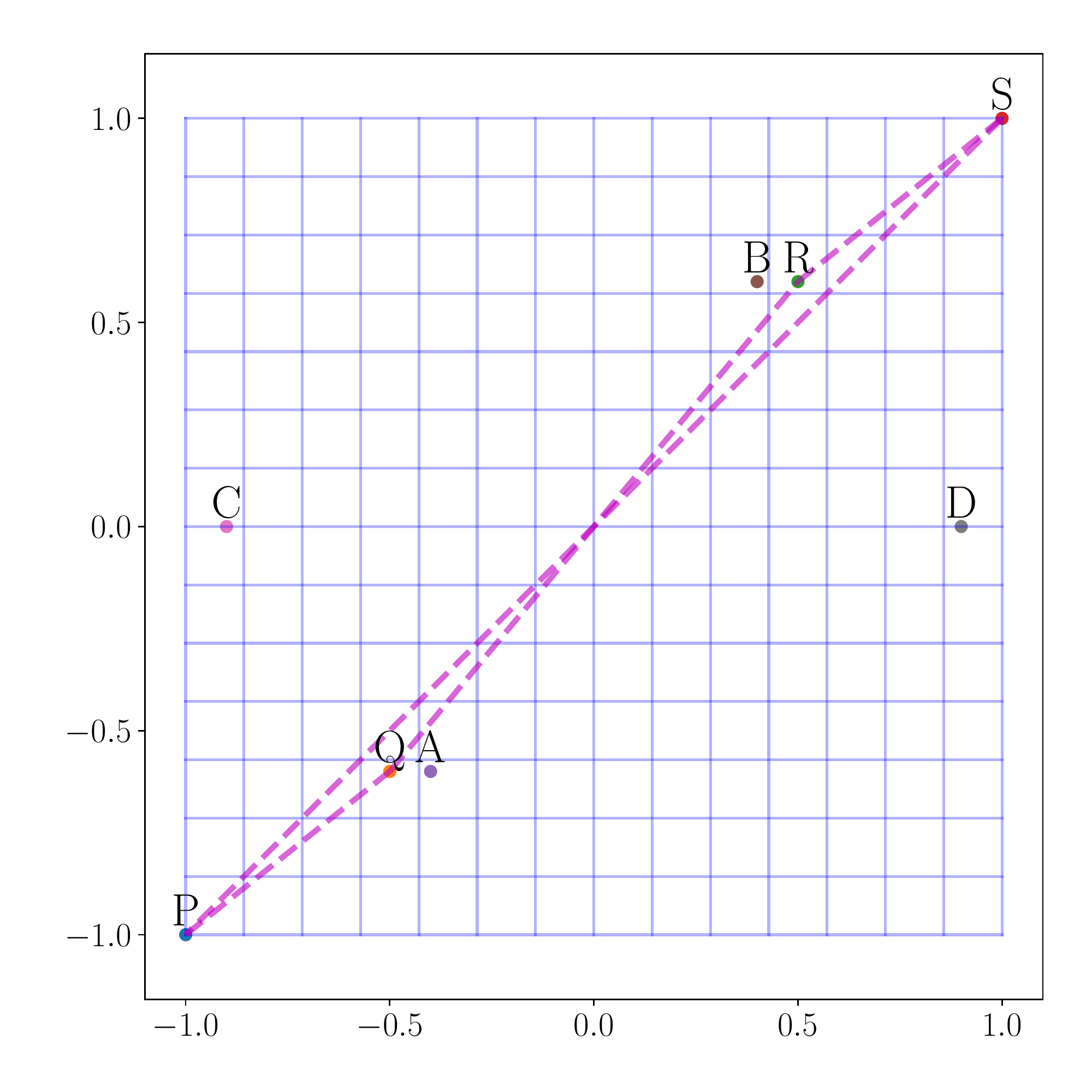}
    \caption{\label{fig:cycle-input} Embeddings and graph}
  \end{subfigure}
  \begin{subfigure}[t]{0.24\textwidth}
    \includegraphics[width=\linewidth]{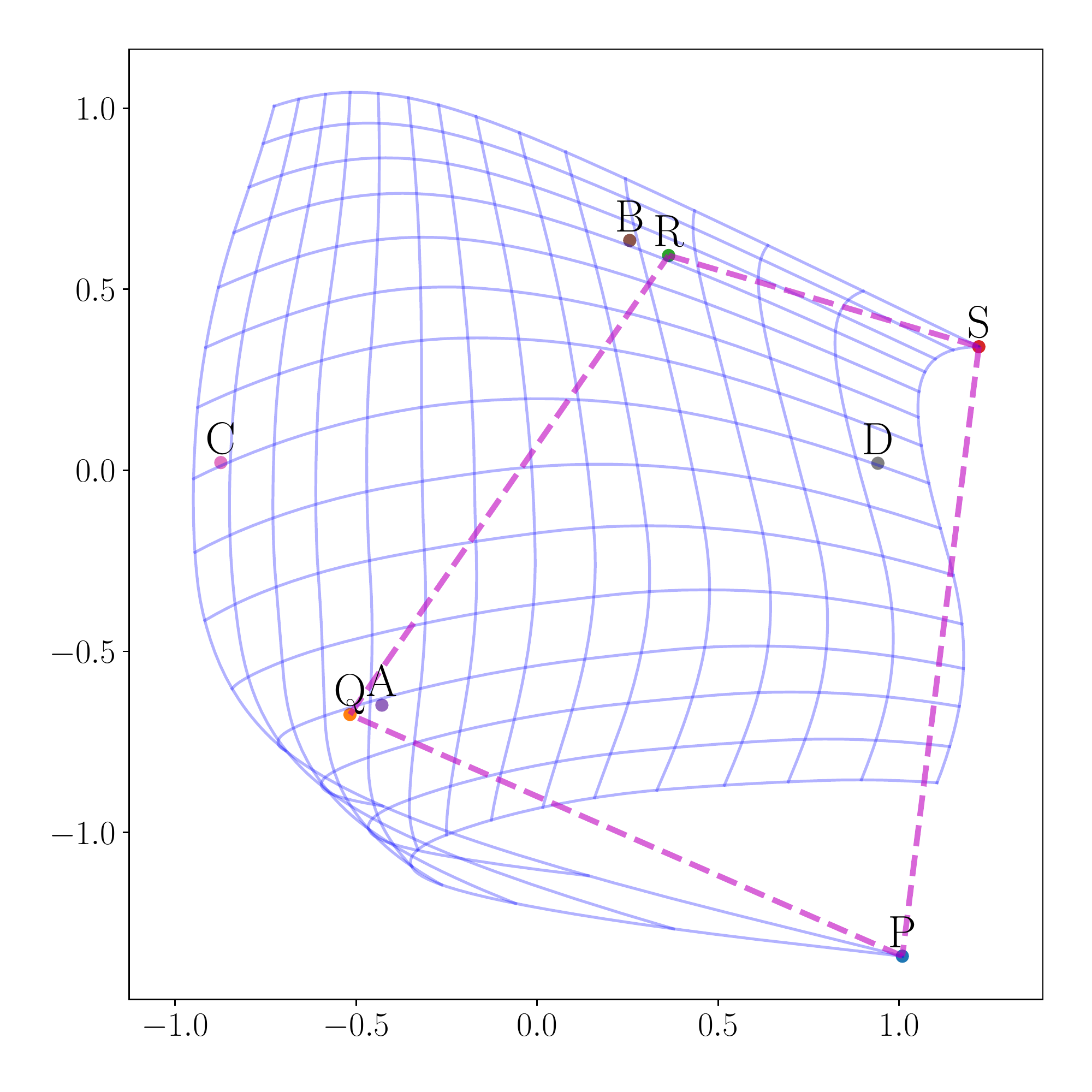}
    \caption{\label{fig:cycle-proximal} Proximal }
  \end{subfigure}
  \begin{subfigure}[t]{0.24\textwidth}
    \includegraphics[width=\linewidth]{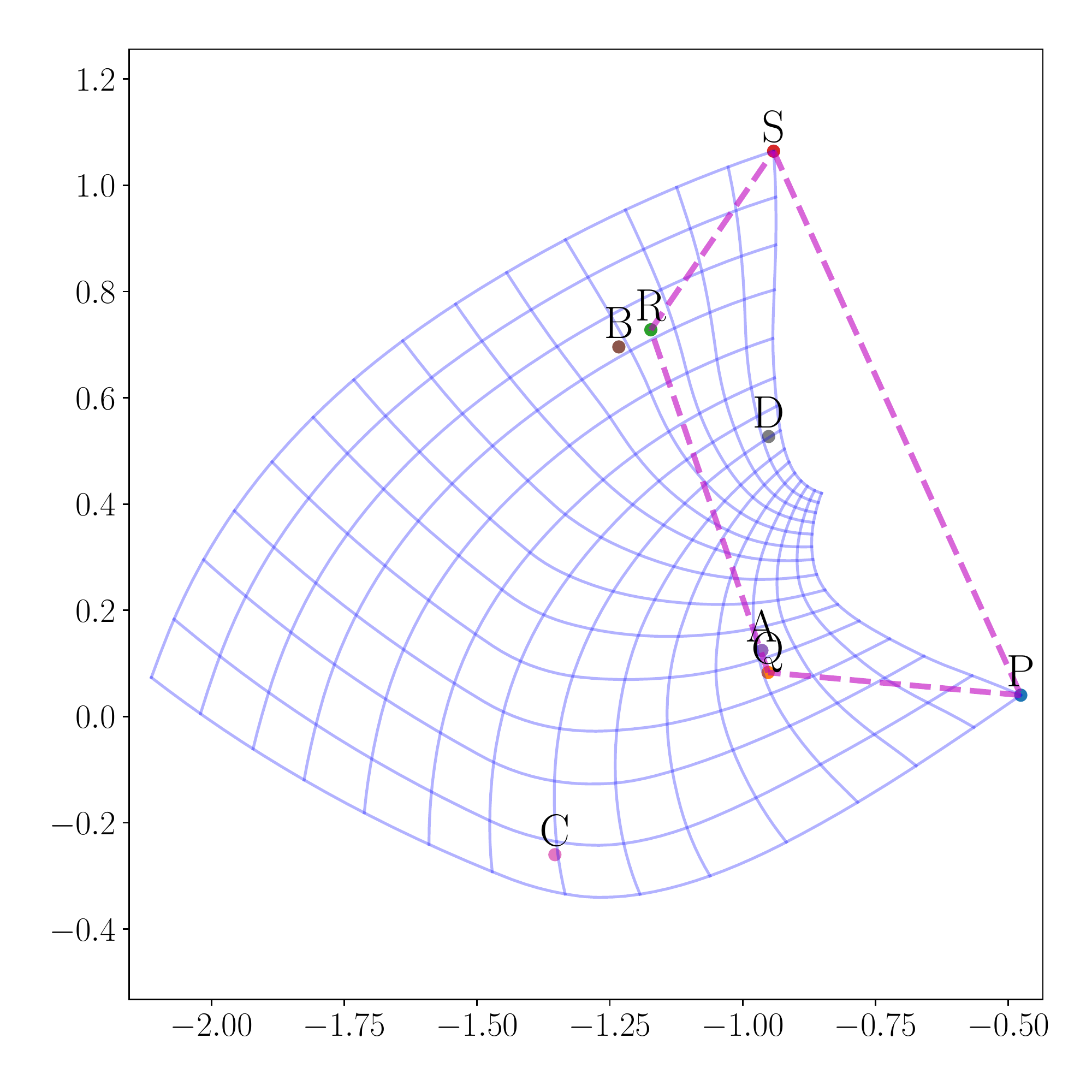}
    \caption{\label{fig:cycle-conformal} Conformal }
  \end{subfigure}
  \begin{subfigure}[t]{0.24\textwidth}
    \includegraphics[width=\linewidth]{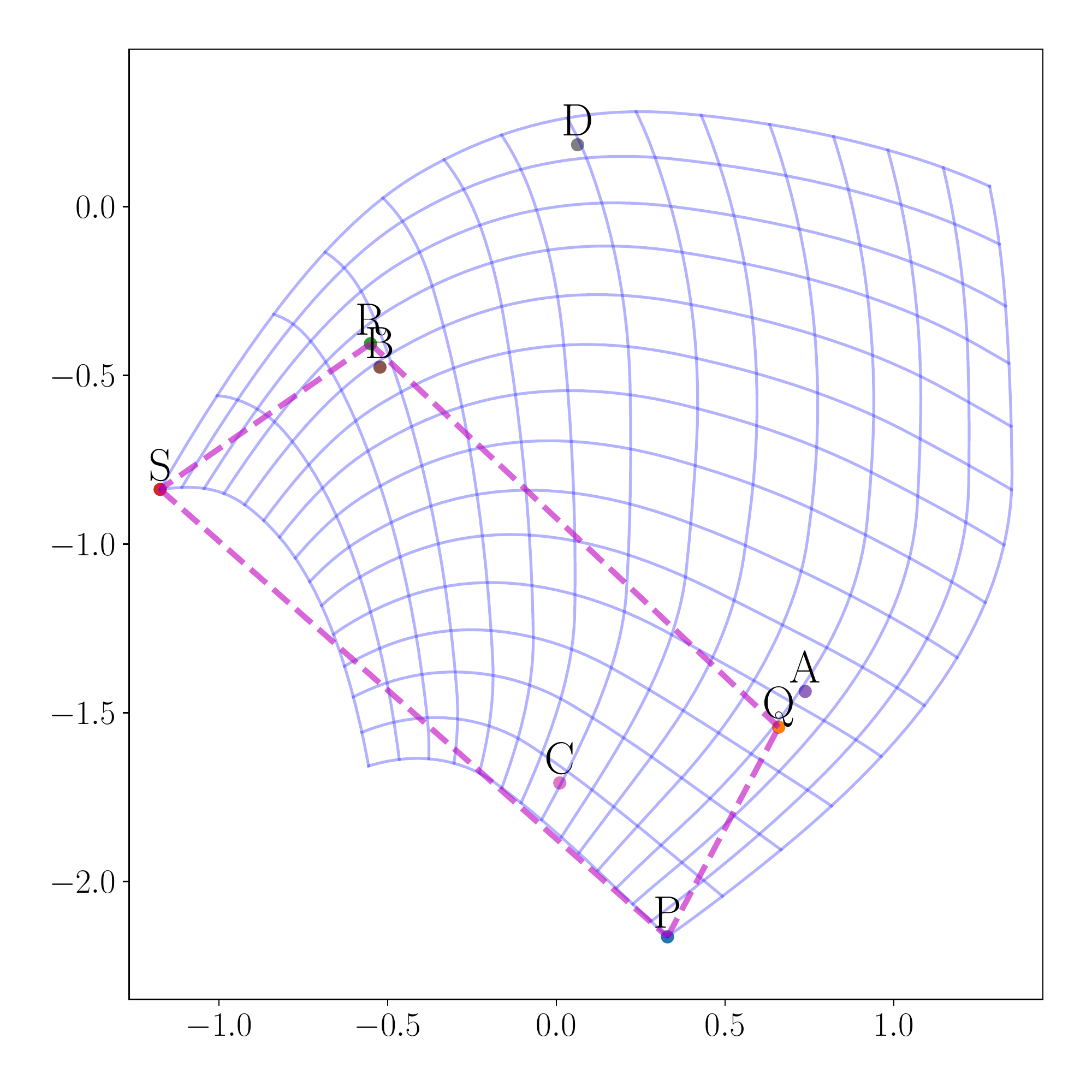}
    \caption{\label{fig:cycle-isometry} Isometric }
  \end{subfigure}
  \caption{\label{fig:cycle} 
  \textit{Comparison of regularization methods for retrofitting.}
  (a) illustrates an example of given embeddings (ABCD and PQRS) and a cycle over the subset PQRS which are approximately collinear in the source embeddings;
    P and S start off far away from each other despite being neighbors in the graph.
  (b) The proximal regularizer used in prior work minimizes how far points in the image move without actually preserving spatial relationships from the input; in particular it learns a map that severely distorts the space between P and Q.
  (c-d) We propose a new \textit{conformality regularizer} that allows points to move arbitrarily, but explicitly preserves spatial relationships such as angles (c) or distances (d) from the input; the learned maps transform space much more smoothly.
  }
\end{figure*}

Downstream methods rely on spatial relationships---distances and angles---between retrofitted embeddings.
The proximity regularizer $\loss_\prox$ used by current retrofitting methods is often unable to preserve these relationships (\reffig{cycle-input}).
In this section, we propose a novel conformality regularizer that explicitly preserves local distances and angles in the image of the map $f$.

For any Riemannian manifold, the local geometry around a point $p \in \mS$ is defined by its tangent space $T_p \mS$, and the Jacobian~\footnote{%
  In differential geometry, $\pushforward{f}_p$ is better known as the \textit{pushforward} or differential.}
  $\pushforward{f}_{p}: T_p \mS \to T_{f(p)} \mT$ describes how $f$ transforms $T_p \mS$.
Using distances between points in this transformed tangent space, $f$ induces a metric---the \textit{pullback metric} $\pullback{f}$---in the source manifold:
 $\pullback{f}_p(x, y) \eqdef g\inT_{f(p)}(\pushforward{f}_p(x), \pushforward{f}_p(y))$ for $x, y \in \mS$.
Written in more familiar matrix notation, the pullback metric tensor at $p$ is:
\begin{align}
  F_p &= {J_p}^\top  G\inT_{f(p)} J_p, \label{eqn:pullback}
\end{align}
where $J_p$ is the Jacobian matrix at $p$, and $G\inT_{f(p)}$ is the target metric tensor at $f(p)$.

When the pullback metric is equal to the original metric at $p$, so are distances and angles in its local geometry.
This motivates an isometry regularizer that minimizes the difference between these metrics:
\begin{align}
  \loss_\isometry(p) &= D\big( F_{p}, G\inS_{p} \big), \label{eqn:iso-reg}
\end{align}
where $D$ is an appropriate distance function, and $G\inS_{p}$ is the source metric tensor. 
In practice, we use the geodesic distance on positive definite matrices $D(X, Y) = \|\log(XY\inv)\|^2$~\cite{lim2019geometric}, a symmetric loss function that is invariant to scalar transformations, congruence transformations and inversion. For example, if both the target metric and the original metric are the identity and representative of Euclidean space, this objective will encourage the Jacobian to be unitary at all points.
We note that the $G_p\inS$ and $G_{f(p)}\inT$ are defined by the given manifolds, and $J_p$ can be easily computed using automatic differentiation.

The isometry regularizer already improves on the proximity regularizer in encouraging smoother maps in \reffig{cycle-isometry}; 
  most notably, it generally preserves the areas of each grid square in the image. 
When angular information is more important than distances, we can relax the equality constraint to one that bounds the ratio between the two metrics:
\begin{align*}
  \loss(p) &= D\big( F_{p}, e^\alpha G\inS_{p} \big) & \text{subj.\ to}~-C \le \alpha \le C,
\end{align*}
where $\alpha$ is a free parameter and $C$ is the desired bound.
In the supplementary material, we solve for KKT conditions and show that when $D$ is the geodesic distance, the above constrained objective reduces to the following unconstrained objective:
\begin{align}
  \loss_\conformal(p) = 
      \left\| \log\left(F_{p} {G\inS_{p}}\inv \right) \right\|^2 - \min\left\{C, {\log\det(F_p {G\inS_p}\inv)}^2 \right\}.
    \label{eqn:conf-reg}
\end{align}
We call $\loss_\conformal$ the conformality regularizer; when $C=0$ it reduces to \refeqn{iso-reg} and preserves distances; when $C \to \infty$ it only preserves angles (\reffig{cycle-conformal}).
Like the geodesic distance, it is invariant to scalar and congruence transformations, as well as matrix inversion.

In the supplementary material, we prove that optimizing this regularizer over the input space is both necessary and sufficient for $f$ to be an isometric and/or conformal map: 
\begin{theorem}
  For all values of $C$, $f$ is a conformal map iff $\loss_\conformal(w) = 0$ for all points $w \in \mS$;
    if $C = 0$, then $f$ is a isometric map iff $\loss_\conformal(w) = 0$ for all points $w \in \mS$.
\end{theorem}

\subsection{\label{sec:rff} Riemannian feed forward layers}
\begin{figure*}[t]
  \begin{subfigure}[t]{0.24\textwidth}
    \includegraphics[width=\linewidth]{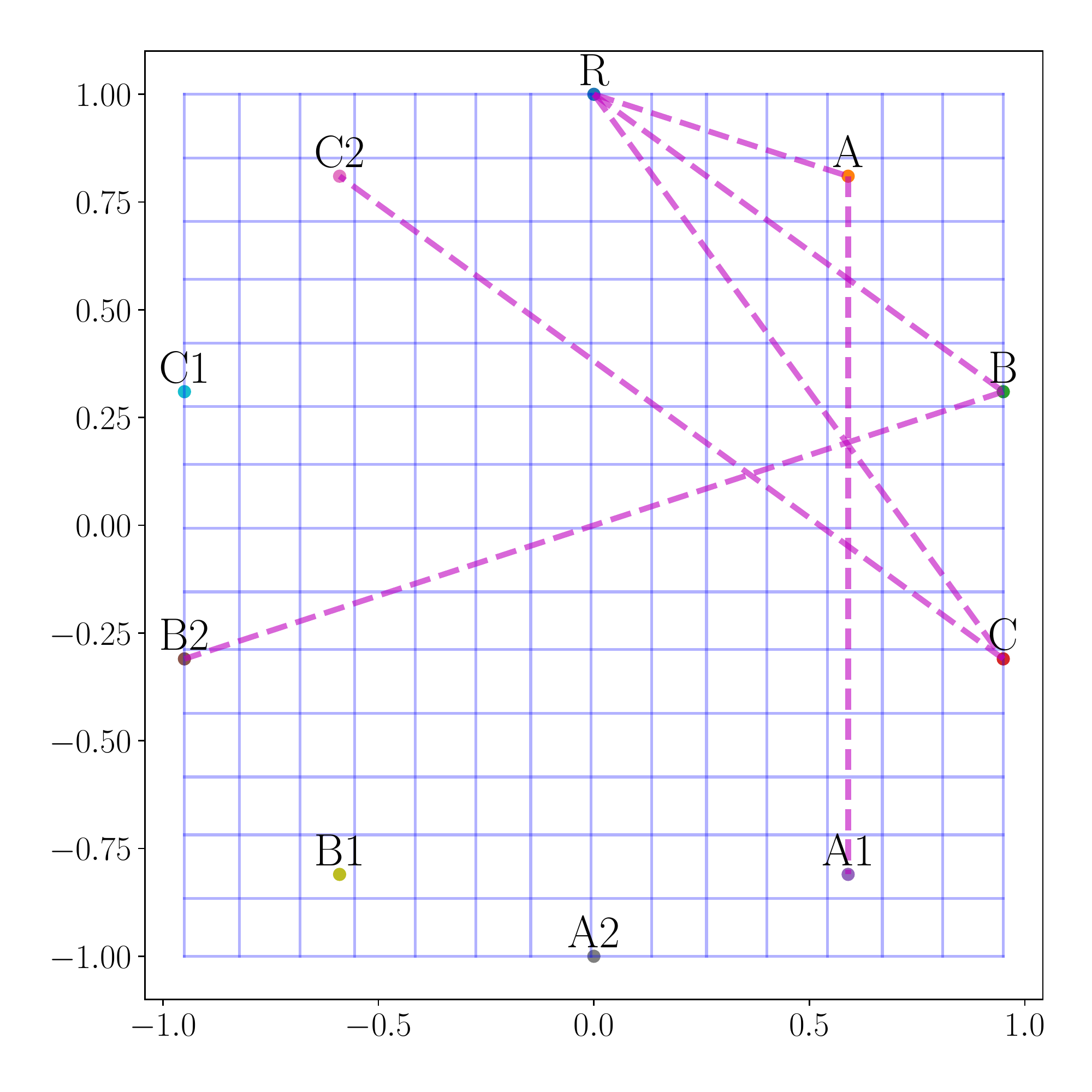}
    \caption{\label{fig:ctree-input} Embeddings and graph}
  \end{subfigure}
  \begin{subfigure}[t]{0.24\textwidth}
    \includegraphics[width=\linewidth]{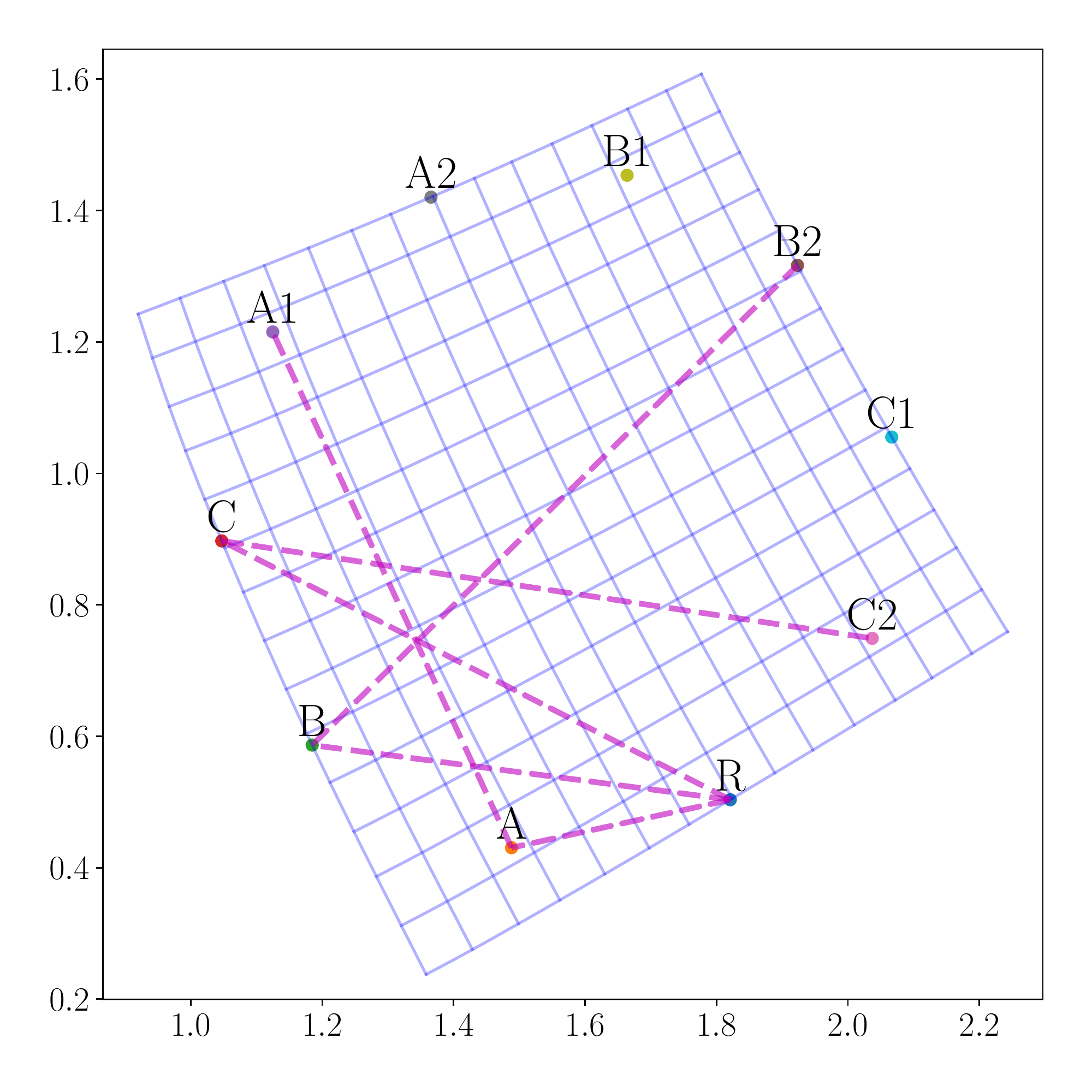}
    \caption{\label{fig:ctree-euc} $\mEuc^2$ }
  \end{subfigure}
  \begin{subfigure}[t]{0.24\textwidth}
    \includegraphics[width=\linewidth]{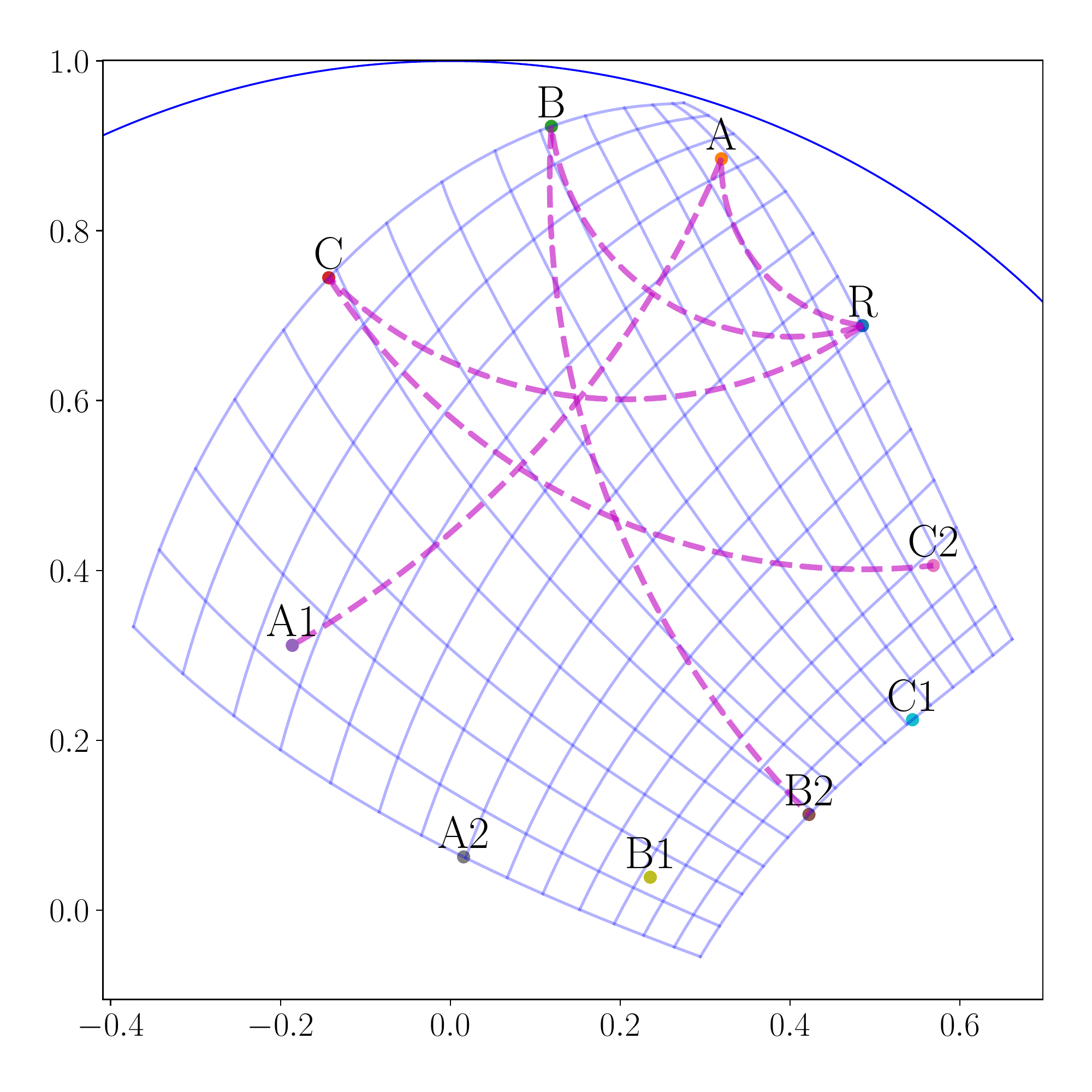}
    \caption{\label{fig:ctree-hyp} $\mHyp^2$ }
  \end{subfigure}
  \begin{subfigure}[t]{0.24\textwidth}
    \includegraphics[width=\linewidth]{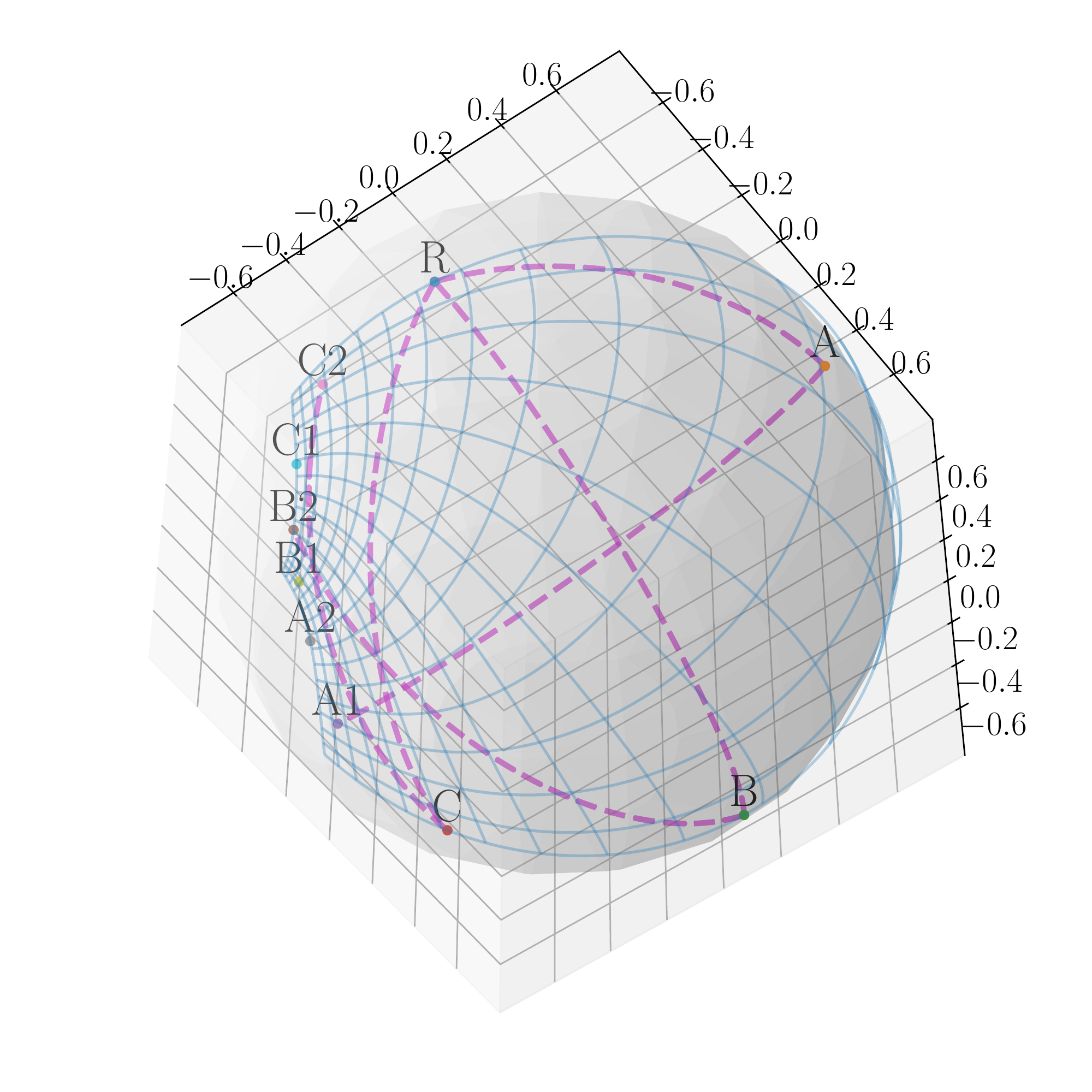}
    \caption{\label{fig:ctree-sph} $\mSph^2$ }
  \end{subfigure}
  \caption{\label{fig:ctree} 
  Comparison of target manifolds when retrofitting given pretrained embeddings (R, A, B, C, A1, A2, B1, B2, C1 and C2) to a tree over the subset R, A, B, C, A1, B1, C2 (a).
  The Euclidean manifold is unable to properly represent the graph without severely distorting space (b).
  The Hyperbolic manifold perfectly represents the graph by exploiting its unique geometry; it embeds points near the edge of the Poincar\'e ball (drawn in blue) where space is sufficiently stretched to ensure that the distance between B and B2 is actually far less than the distance between (say) B and C (c).
  The Spherical manifold perfectly represents the graph by exploiting its geometry; A is able to be close to A1, A2 (similarly for B and C) by following a path on the far side of the sphere (d).
  }
\end{figure*}

So far, our focus has been on learning Euclidean target embeddings.
However, for many graphs Euclidean manifolds can require far more dimensions than non-Euclidean manifolds~\cite{matouvsek1999embedding,nickel2017poincare,gu2019learning};
  this problem is only exacerbated if spatial relationships from the input must be maintained.
\reffig{ctree} provides one such example where non-Euclidean manifolds, unlike the Euclidean one, are able to exploit their unique geometry to fit a graph while maintaining spatial relationships.
We generalize standard feedforward layers to accept and project embeddings on general Riemannian manifolds.

\providecommand{\sigmoid}{\sigma}

A typical (Euclidean) feedforward layer ($\EFL{}$) transforms input $x \in \Re^m$ by applying a linear transform $A \in \Re^{n \times m}$, followed by a translation by $b \in \Re^n$ and a pointwise nonlinearity $\sigmoid$:
$\EFL(x) \eqdef \sigmoid \left( A x + b \right)$.
However, none of these steps have direct analogs in Riemannian geometry.
We overcome this limitation by applying the transforms in Euclidean tangent spaces using the $\log$ and $\exp$ maps;
the Riemannian feedforward layer ($\RFL{}$) from source manifold $\mS$ and target manifold $\mT$ is:
\begin{align}
  \RFL(x\inS) &\eqdef \exp_{b\inT} \left(
      \sigmoid \left( A \log_{b\inS} x\inS \right)
      \right), \label{eqn:rfl}
\end{align}
where $b\inS \in \mS$, $b\inT \in \mT$ are distinct bias terms and $A: T_{b\inS} \mS \to T_{b\inT} \mT$ is a linear operator between the source and target tangent spaces.
For the manifolds considered in this paper, which are \textit{geodesically complete}, we can represent $A$ as a matrix in $\Re^{n \times m}$,
  where $m$ and $n$ are respectively the dimensions of the source and target manifolds.
When applied to Euclidean manifolds, $\RFL{}$ reduces to $\EFL$ when $b\inT = 0$ and $b\inS = -A\pinv b$;
  while the two bias terms are redundant in this case, they are necessary for Riemannian manifolds that do not contain a $\bzero$ element like $\mSph$.
Derivatives for all of these parameters can be efficiently computed through automatic differentiation.

Like typical feedforward layers, two Riemannian feedforward layer can be stacked when the output manifold of the first layer is the input manifold of the second.
We call such stacked Riemannian feedforward layers \textit{Riemannian feedforward networks}.

\subsection{Putting it together: conformal retrofitting} 
The final component of a retrofitting method is the fidelity loss $\loss_\fidelity$ used to encourage graph neighbors to be closer to each other in the target embeddings.
We found that a manifold-aware max-margin loss worked best:
\begin{align}
  \loss_\fidelity^\cretro(u, v) &= \sum_{x\inT \in \sN(u\inT)} \max(0, \gamma + d(u\inT, v\inT) - d(u\inT, x\inT)), \label{eqn:cretro}
\end{align}
where $\gamma$ is the margin, $\sN(u\inT)$ is the set of neighbors to $u\inT$ in the target manifold excluding any graph neighbors;
  the distances, $d(u\inT, v\inT)$ and $d(u\inT, x\inT)$, are also measured in the target manifold.

Putting all these pieces together, we propose a new retrofitting method, \textit{conformal retrofitting}, that transforms pretrained embeddings using a Riemannian feedforward network, $u\inT = f_\theta(u\inS)$ and is trained with the following objective:
\begin{align}
  \sL^{\cretro} &= 
  \sum_{(u, v) \in E} \loss_\fidelity^\cretro(u, v) +
    \lambda \sum_{w \in V} \underbrace{\loss_\conformal(w\inS)}_{\loss_\preservation^\cretro(w)}. \label{eqn:cretro-full}
\end{align}

\section{\label{sec:experiments} Experiments}

\subsection{Training details}

We use Riemannian Stochastic Gradient Descent (R-SGD)~\cite{bonnabel2013riemannian}---an extension of stochastic gradient descent that efficiently projects gradient updates from the tangent space onto the manifold---to train parameters of non-Euclidean Riemannian feedforward layers, and Adam~\cite{kingma2014adam} to train the remaining Euclidean parameters.
We found that the relative scales of the fidelity and preservation losses changed significantly over the course of training and a static objective weight $\lambda$ would only train one of the two losses.
To solve this problem, we used GradNorm~\cite{chen2018gradnorm}, an adaptive loss balancing algorithm that weighs objectives inversely proportional to norm of their gradients;
  we found it necessary to modify the algorithm to use the geometric mean of gradient norms instead of the arithmetic mean to be more robust to outliers.

The max-margin loss $\loss_\fidelity^\cretro$ uses nearest neighbors in $V\inT$. During training, we sampled 50 manifold neighbors for each point in a mini-batch.
While a number of fast exact and approximate near neighbor algorithms exist for Euclidean embeddings,
  they rely on fast distance and mean computations.
Both of these operations can be significantly slower for non-Euclidean manifolds, even those with closed-form distance functions like $\mHyp$ or $\mSph$.
Following \citet{turaga2010nearest},
  we overcome this bottleneck by projecting $V\inT$ onto the tangent space at their centroid:
  $\arg\min_{c \in \mT} \sum_{v\inT \in V\inT} d(c, v\inT)^2$.
We then build an efficient nearest neighbor index over these Euclidean projections using the FAISS library~\cite{JDH17};
  the index is periodically updated as the model trains.

\subsection{Evaluation setup}

For all experiments below, we used 
    50-dimensional pretrained GloVe embeddings~\citep{pennington2014glove} as our source embeddings,
    two Euclidean intermediate layers in $\mEuc^{1600}$ and
    vary the target manifold.\footnote{The dimensionality of the intermediate layers were chosen after an initial random grid search.}
We restricted intermediate layers for all methods to be Euclidean to fairly compare with explicit retrofitting,
  and to focus our hyper-parameter search on the target manifold.
When picking the search space for the target manifold, 
we focused on two settings: (a) where the total dimensions were equal to the source embeddings (50-dimensional) to compare with baselines, and (b) where they were slightly larger (60-dimensional) to explore the benefits of added dimensions.
In each setting, we explored both pure manifolds (e.g. $\mathbb{S}^{50}$) and product manifolds that either were a balanced or skewed split. 

Additionally, for each target manifold, we tuned conformality and learning rate using random grid search.
The conformality parameter ($-\log(C)$) was chosen using uniformly spaced values from its entire range: $0, 0.2, 0.4, 0.6, 0.8, 1$.
Finally, we ran each hyperparameter combination once and report the results of the best model (using validation metrics) in each hyperparameter sweep. 
In our final experiments, we selected the baseline model (explicit retrofitting) from a sweep of 30 runs for each dataset,
and the proposed conformal retrofitting model from a sweep (that included the target manifold and conformality as hyperparameters) of 60 runs for each dataset.
All results are presented using early stopping on the validation set.

We evaluate the above retrofitting methods using 
  two datasets constructed from the WordNet~\cite{miller1995wordnet} graph:
  the hierarchy of all 1,180 mammals connected via a hypernymy relation (\DMammals), and  
  the hierarchy of all 82,061 nouns connected via a hypernymy relation (\DNouns).
The nodes of datasets were split into train, validation and test sets using a $80::10::10$ split.
While the train set only contains edges between its nodes, the validation set includes edges to the train set and the test set includes edges to both train and validation sets.

Similar to \citet{glavas2018explicit}, we measure similarity using cosine distance, which is equivalent to being embedded on $\mSph$.
\citet{glavas2018explicit} use a slightly different contrastive loss and method to sample near neighbors.
To fairly compare our methods, we reimplement explicit retrofitting using our max margin loss and neighborhood sampling with the proximal regularizer.
We use mean average precision (mAP) to evaluate each methods ability to predict hypernymy relations (edges) to words (nodes) not seen during training.
We also report scores on metrics achieved by the original GloVe embeddings.

Additional details, dataset statistics and a complete list of chosen hyperparameters are provided in the supplementary material.

\begin{table*}[t]
  \centering
  \begin{subfigure}[t]{0.49\textwidth}
    \centering
\begin{tabular}{lrrr}
\toprule
                           & & Train       & Test \\
\multicolumn{2}{c}{Retrofitting Method}   & mAP         & mAP  \\
\midrule
\multicolumn{2}{l}{Original ($\mSph^{50}$)}
  & 9.9\%   & 7.9\% \\
\multicolumn{2}{l}{Explicit ($\mSph^{50}$)}
  & 11.5\%  & 8.9\% \\
\midrule
Conformal & $-\log(C)$ \\
\cmidrule{1-2}
  $\mSph^{50} $ & 1.0    
  & 11.6\%  & 9.0\% \\
  $\mSph^{30}\times\mEuc^{30}$ & 0.2 
  & 11.9\%  & 9.2\% \\
  $\mSph^{60}$ & 0.8
  & 11.9\%  & \textbf{9.3\%} \\
  $\mSph^{30}\times\mHyp^{30}$ & 0.2  
  & \textbf{12.8\%}  & \textbf{9.3\%} \\ 
  \bottomrule
\end{tabular}

    \caption{\label{tab:nouns} \DNouns}
  \end{subfigure}
  \begin{subfigure}[t]{0.49\textwidth}
    \centering
\begin{tabular}{lrrr}
\toprule
                           & & Train       & Test \\
\multicolumn{2}{c}{Retrofitting Method}   & mAP         & mAP  \\
\midrule
\multicolumn{2}{l}{Original ($\mSph^{50}$)}
  & 25.1\% & 21.5\% \\
\multicolumn{2}{l}{Explicit ($\mSph^{50}$)}
  & \textbf{73.4\%} & \textbf{26.3\%} \\
\midrule
Conformal & $-\log(C)$ \\ \cmidrule{1-2}
$\mSph^{50}\times\mEuc^{5}\times\mHyp^{5}$ & 0.6
& 69.4\% & 25.6\% \\
$\mSph^{30}\times\mEuc^{30}$ & 0.4
& 45.2\% & 26.0\% \\
$\mSph^{45}\times\mEuc^{5}$ & 0.6
& 34.9\% & 26.1\% \\
$\mSph^{50}\times\mEuc^{10}$ & 0.2
  & 52.9\% & \textbf{26.3\%} \\
\bottomrule
\end{tabular}

    \caption{\label{tab:mammals} \DMammals}
  \end{subfigure}
  \caption{\label{tab:evaluation} 
  \textit{Link prediction scores measured using mean average precision (mAP) on two hypernymy datasets (\DNouns{} and \DMammals{}).} 
  We report conformality as $-\log(C)$, with a value of $0$ corresponding to $C=\infty$, to better represent its range.
  On \DNouns, conformal retrofitting improves test mAP scores over explicit retrofitting even in the original $\mSph^{50}$ manifold;
    the best results are obtained in higher dimensions.
  On both datasets, mixed non-Euclidean manifolds significantly improve train mAP scores, reflecting their better ability to represent the graph seen during training.
  }
\end{table*}

\begin{figure}[t]
  \centering
  \begin{subfigure}[t]{0.49\textwidth}
    \centering
    \includegraphics[width=\linewidth]{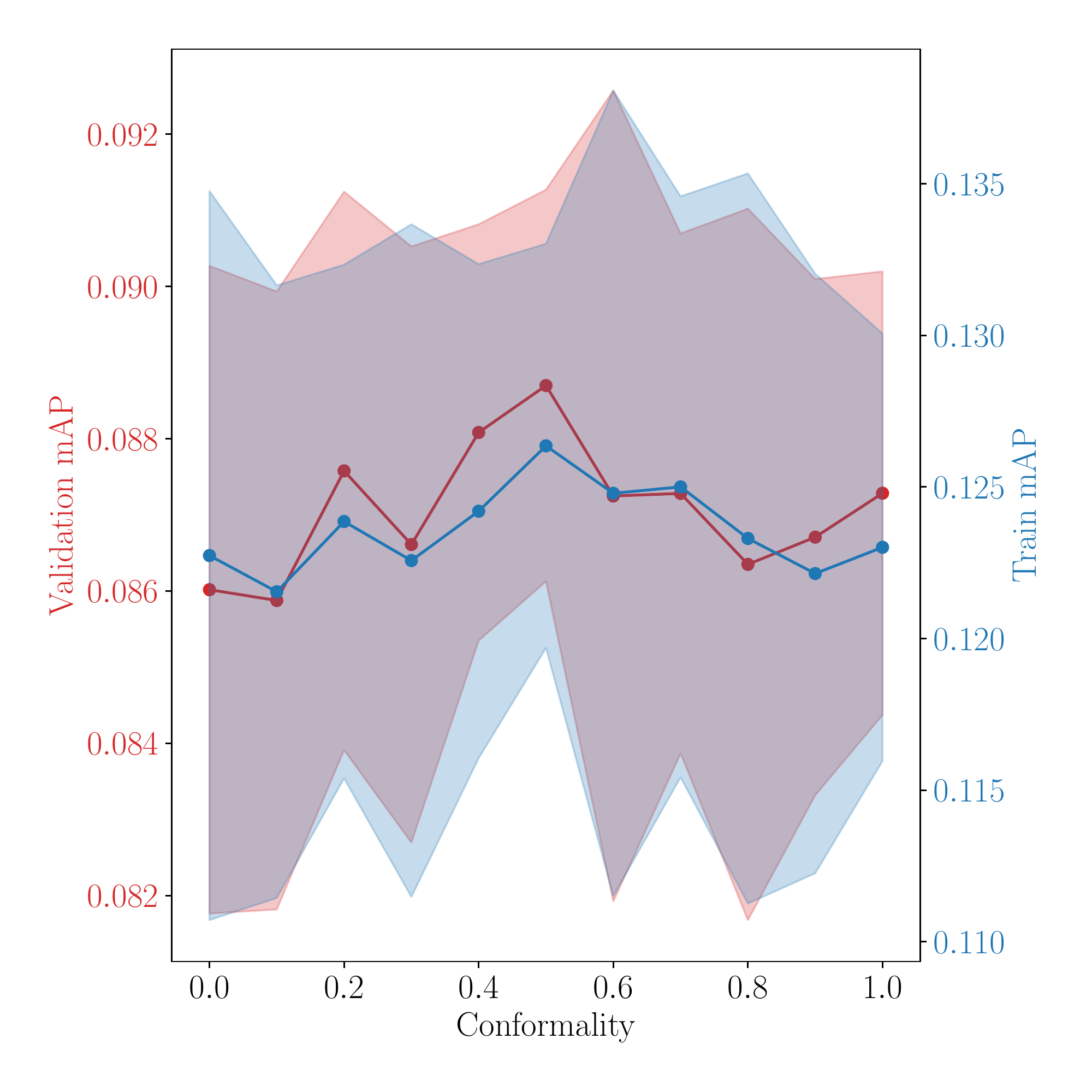}
    \caption{\label{fig:conformality-nouns} $\mSph^{30}\times\mHyp^{30}$ on \DNouns}
  \end{subfigure}
  \begin{subfigure}[t]{0.49\textwidth}
    \centering
    \includegraphics[width=\linewidth]{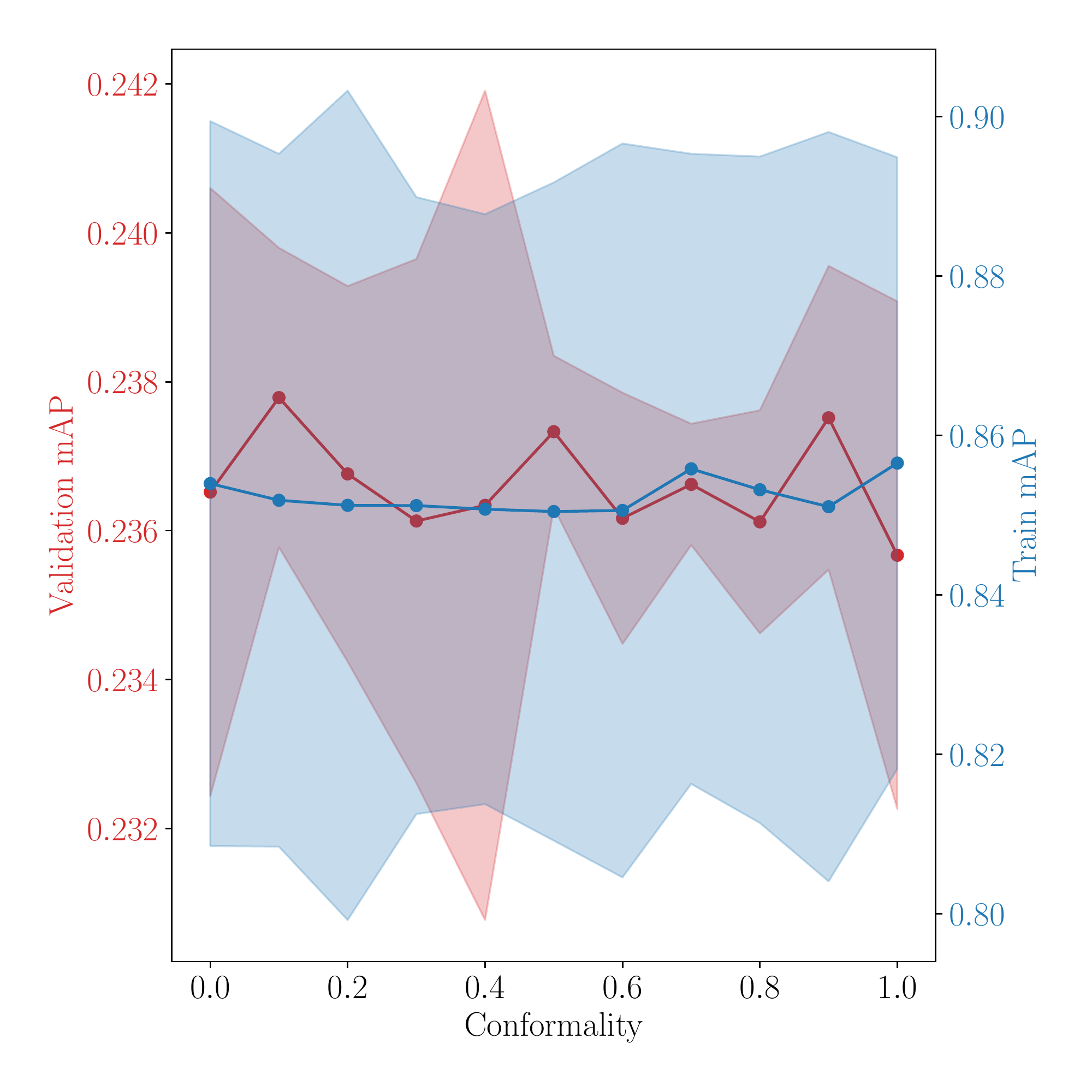}
    \caption{\label{fig:conformality-mammals} $\mSph^{50}\times\mEuc^{10}$ on \DMammals}
  \end{subfigure}
  \caption{\label{fig:conformality} \textit{Impact of conformality on train and validation mAP scores for top-performing target manifolds.}
  Each point is the average over 5 runs; the shaded region highlights a single standard deviation.
  We report conformality as $-\log(C)$, with a value of $0$ corresponding to $C=\infty$, to better represent its range.
  For $\mSph^{30}\times\mHyp^{30}$ on $\DNouns$, we find that an intermediate values perform best on both train and validation mean average precision (mAP);
  for $\mSph^{50}\times\mEuc^{10}$ on $\DMammals$, conformality plays less of a role.
  In the appendix, we show that conformality plays a significant role for other target manifolds on $\DMammals$.
  }
\end{figure}

\subsection{Results}

On \DNouns, conformal retrofitting significantly improves test mAP scores when targeting higher dimensional manifolds~\reftab{evaluation}, while still providing improvements in source manifold $\mSph^{50}$.
While purely spherical manifolds ($\mSph^{60}$) have similar test link prediction scores as mixed manifolds ($\mSph^{30}\times\mHyp^{30}$) the latter is significantly better at representing the train graph.
On \DMammals, conformal retrofitting does not provide significant improvements over explicit retrofitting; this is likely due to the small size of the dataset.
However, because \DMammals{} is a more structured graph, we found that mixed manifolds did significant better than their pure counter parts, both on train and test mAP scores.
The conformality hyper-parameter values lie in-between the two extremes, indicating that while slight distortions to distances helps the model fit the graph better, they remain important (\reffig{conformality}).

\section{\label{sec:related} Related Work}

There are several alternatives to retrofitting when combining task-specific graphs with distributional data:
\citet{wang2018zeroshot,peng2019fewshot,chen2019knowledge} encode the task-specific graph as a graph convolutional network that transforms pretrained word embeddings into (visual) object classifiers for unknown or rare labels;
\citet{kumar2016incorporating} incorporate relational constraints from the graph as an additional subspace constraint when learning word-vectors;
\citet{lauscher2019informing} introduce a new pre-training task to predict relations from the graph for contextual embedding models like BERT~\cite{devlin2019bert}.

The topic of better representing graph structures has been well studied:
\citet{mrksic2017semantic,glavas2019monolingual,rothe2017autoextend} extend similarity-based retrofitting~\cite{faruqui2015retrofitting} to include antonymy and directional lexical entailment relations through relation-specific loss objectives complimentary to our own;
\citet{nickel2017poincare} show that hyperbolic manifolds could better represent tree-structured graphs, while \citet{gu2019learning} show that the mixed product-manifolds studied in this paper can better represent more complex graphs in low dimensions;
\citet{balazevic2019multi} apply hyperbolic manifolds to multi-relational graphs.

Riemannian feed-forward layers extend hyperbolic neural networks~\cite{ganea2018hyperbolic}, which are explicitly parameterized for hyperbolic manifolds, to arbitrary Riemannian manifolds. To the best of our knowledge, ours is the first work to define fully-differentiable layers between arbitrary Riemannian manifolds.

\section{\label{sec:conclusion} Conclusions}
%

In this paper, we introduce a new retrofitting method, conformal retrofitting, that can successfully combine task-agnostic representations with graph-structured, task-specific information to produce powerful pretrained embeddings that can be effectively utilized by downstream tasks in natural language and vision.
Specifically, our method comprises of two novel components that we develop:
  (i) a conformality regularizer using the pullback metric from Riemannian geometry, which explicitly encourages the map to preserve angles and distances; and
  (ii) a new neural network layer (the Riemannian feedforward layer) that can learn mappings to non-Euclidean manifolds (to faithfully represent graph structure). 
This enables conformal retrofitting to address key limitations of existing retrofitting algorithms.
We demonstrate the efficacy of conformal retrofitting through experiments on synthetic data with known ground truth structure and on WordNet where conformal retrofitting outperforms existing algorithms by  learning embeddings in non-Euclidean product manifolds.
Our contributions provide an important foundation for future work on both task-specific embeddings, and performance improvements on new downstream applications.

\section*{Broader Impact}

This work is primarily focused on introducing fundamental algorithms and core analysis for pretrained embeddings.
We believe better such algorithms may be helpful for reducing the compute, energy and carbon footprints of developing ML models due to more effective representations and feature reuse.
However, all such algorithms must be trained on well curated, and representative datasets, and should be thoroughly validated to detect potential biases.


\bibliography{all,new}
\bibliographystyle{abbrvnat}

\appendix


\section{Supplementary material: Conformal retrofitting via Riemannian manifolds}

\subsection{Additional Dataset Statistics} 
Below are the additional dataset statistics for $\DNouns$ and $\DMammals$ referenced in the main paper.
\begin{center}
\begin{tabular}{lrrr}
\toprule
                & Nodes & Edges & Mean Degree \\
\midrule
Mammals 
  &  944 / 118 / 118
  & 762 / 234 / 184
  & 1.6 / 2.1 / 1.6 \\
Nouns 
& 65639 / 8211  / 8211
& 53572 / 14700 / 16155 
& 1.6 / 1.9 / 2.1 \\
\bottomrule
\end{tabular}
\end{center}

\begin{figure}[t]
  \centering
  \begin{subfigure}[t]{0.49\textwidth}
    \centering
    \includegraphics[width=\linewidth]{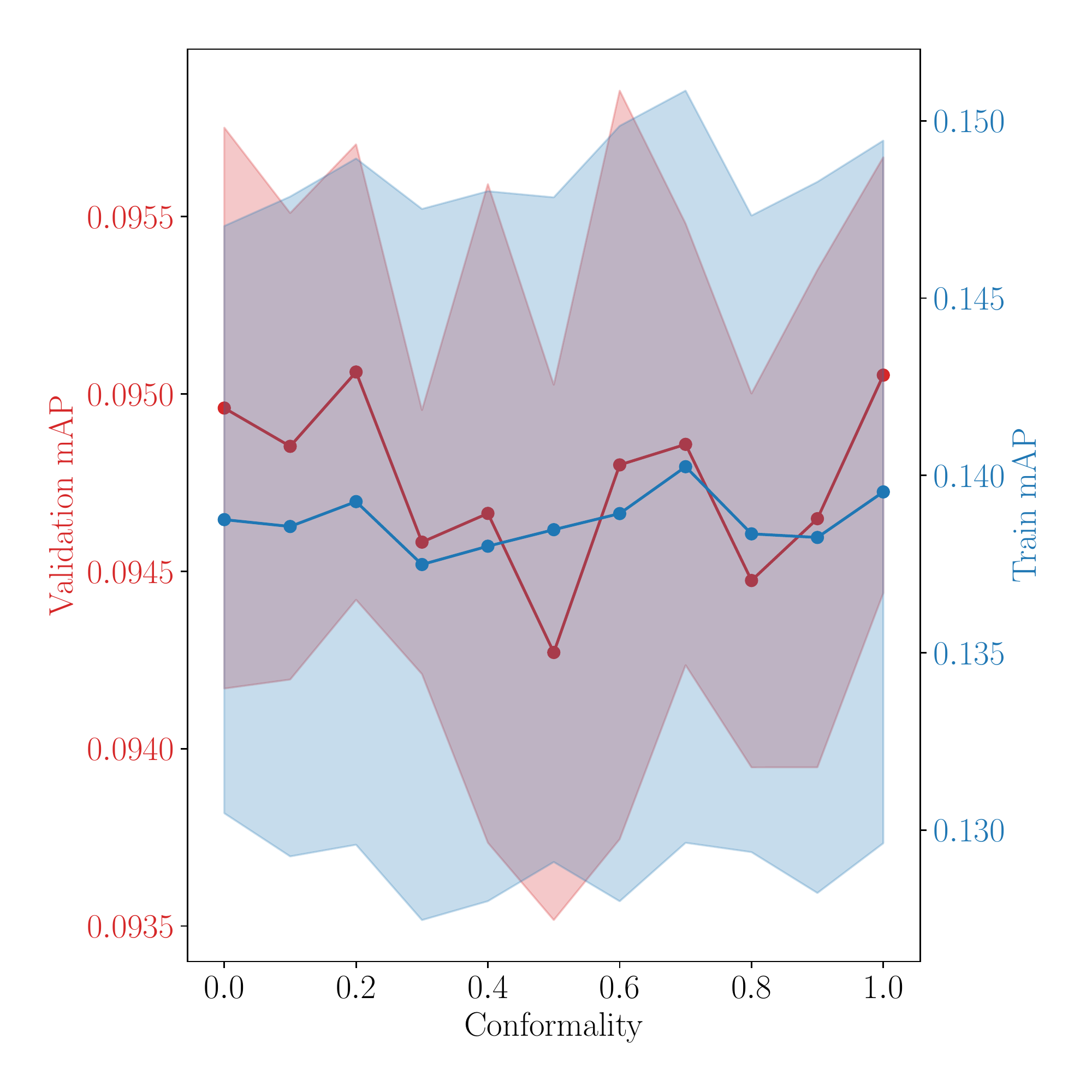}
    \caption{\label{fig:conformality-nouns-s60} $\mSph^{60}$ on \DNouns}
  \end{subfigure}
  \begin{subfigure}[t]{0.49\textwidth}
    \centering
    \includegraphics[width=\linewidth]{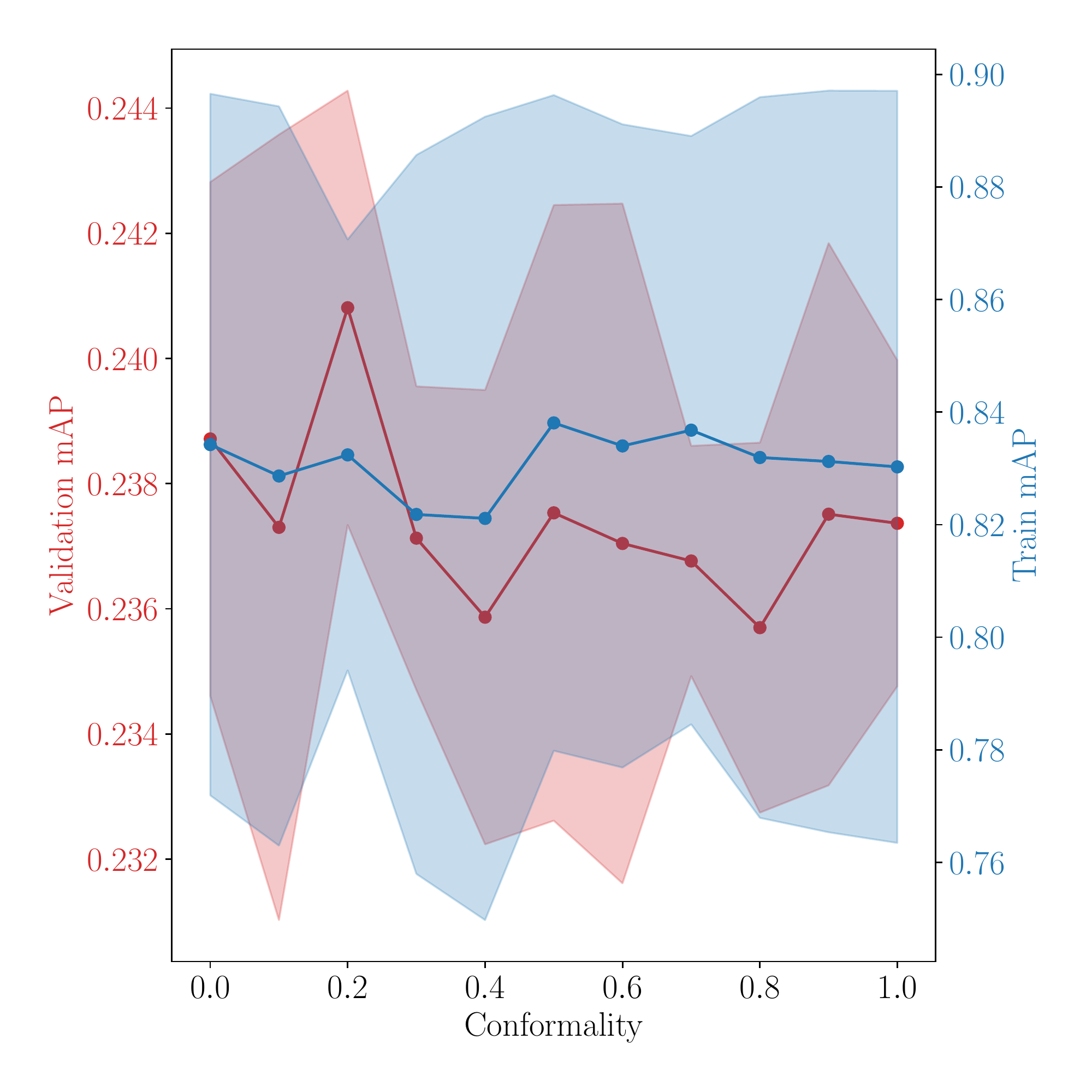}
    \caption{\label{fig:conformality-mammals-s30-e30} $\mSph^{30}\times\mEuc^{30}$ on \DMammals}
  \end{subfigure}
  \\
  \begin{subfigure}[t]{0.49\textwidth}
    \centering
    \includegraphics[width=\linewidth]{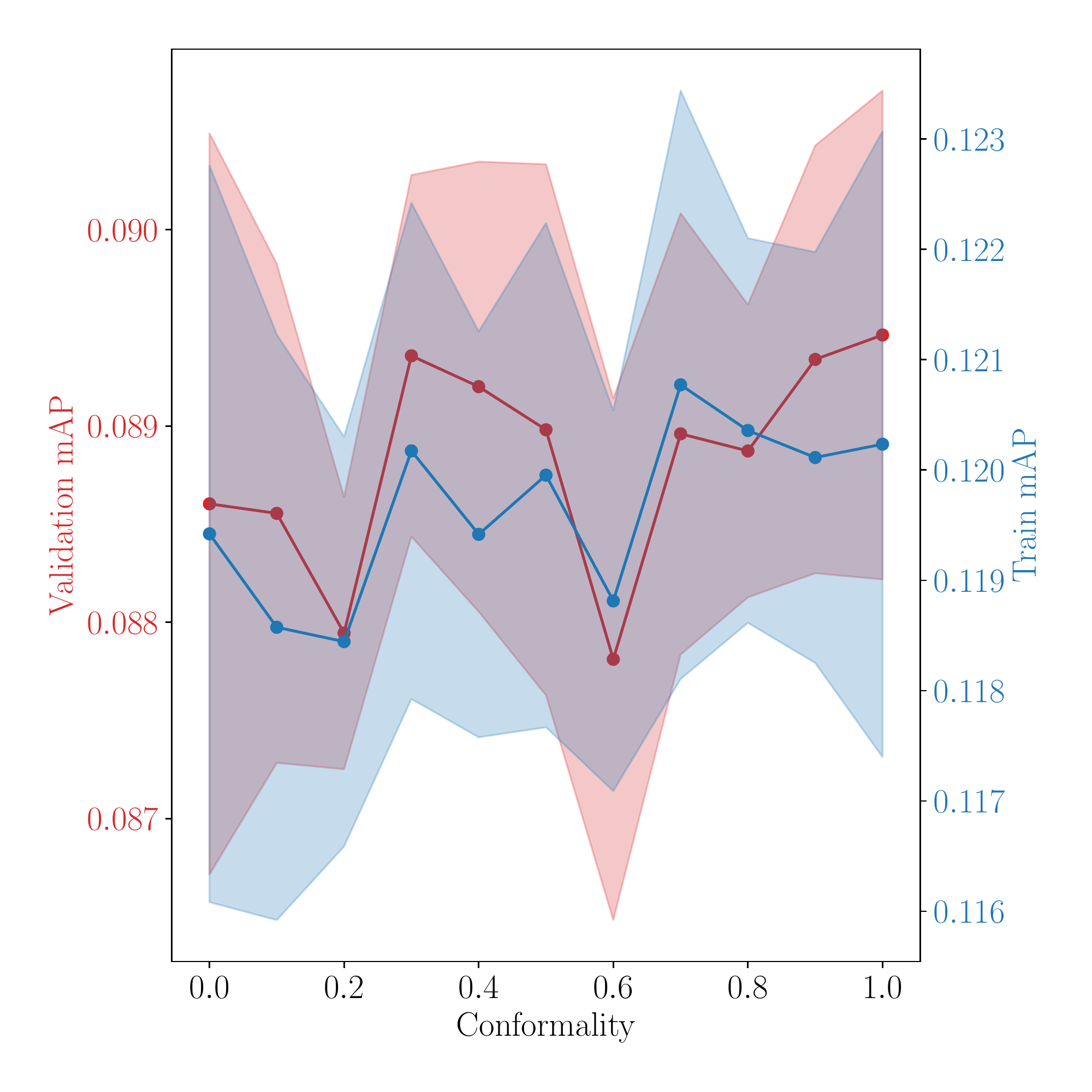}
    \caption{\label{fig:conformality-nouns-s30-e30} $\mSph^{30} \times \mEuc^{30}$ on \DNouns}
  \end{subfigure}
  \begin{subfigure}[t]{0.49\textwidth}
    \centering
    \includegraphics[width=\linewidth]{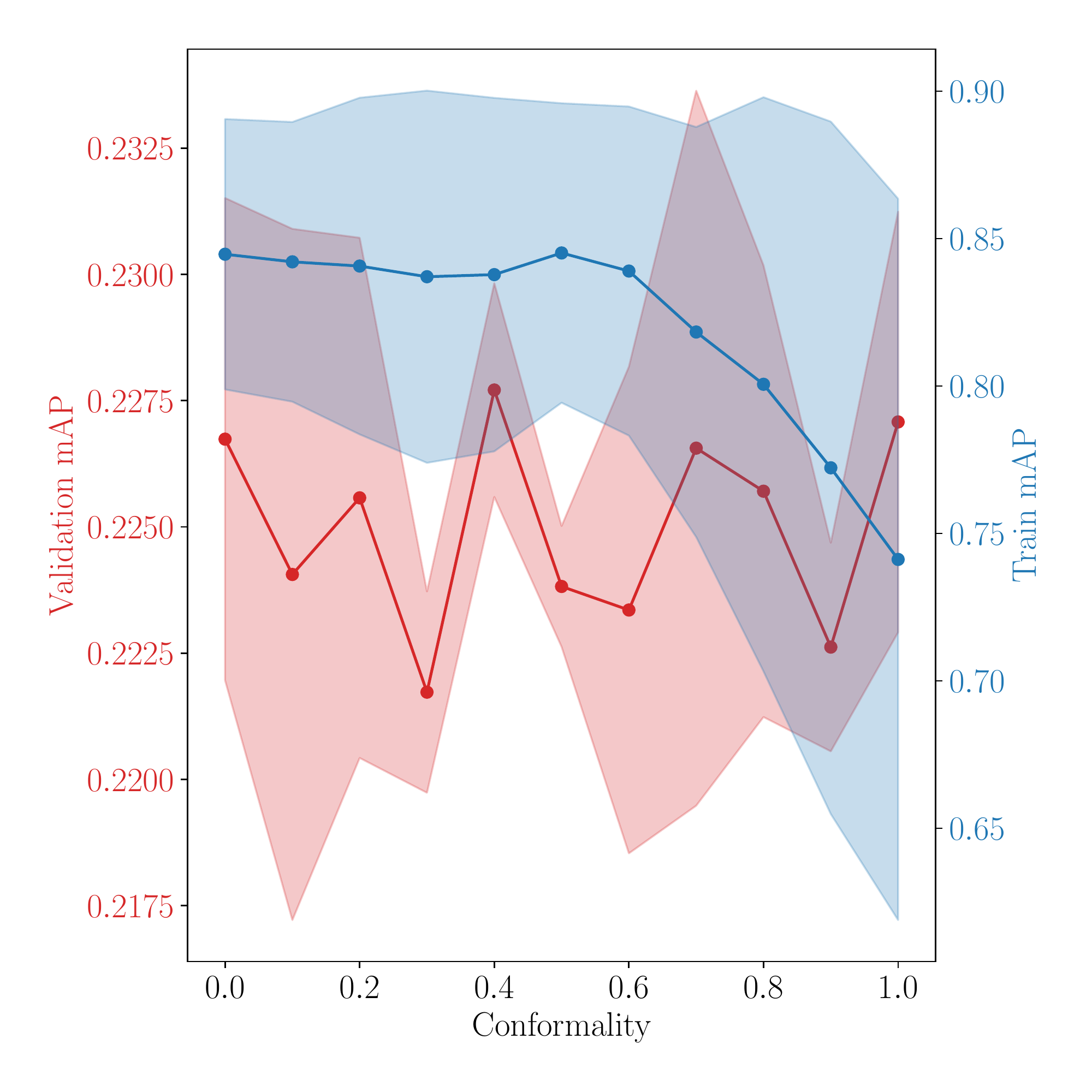}
    \caption{\label{fig:conformality-mammals-s50-e5-h5} $\mSph^{50}\times\mEuc^{5}\times\mHyp^{5}$ on \DMammals}
  \end{subfigure} \\
  \caption{\label{fig:conformality-additional}
  \textit{Impact of conformality on train and validation mAP scores for top-performing target manifolds.}
  Each point is the average over 5 runs; the shaded region highlights a single standard deviation.
  We report conformality as $-\log(C)$, with a value of $0$ corresponding to $C=\infty$, to better represent its range.
  Overall, we find that conformality plays a larger role for mixed manifolds, particularly for those containing a hyperbolic component.
  }
\end{figure}

\subsection{Hyperparameter Details}
To discover the best hyperparameters for our models we performed random searches over several of them. We explored models with one to four Euclidean intermediate layers and layer sizes ranging from 100 to 1600. In a first pass, we found a layer size of 1600 and 3 hidden layers to work best; we fixed these parameters for subsequent experiments. We searched over several different target manifolds: $\mSph^{60}$, $\mHyp^{60}$, $\mEuc^{60}$, $\mSph^{50}$, $\mHyp^{50}$ $\mEuc^{50}$, $\mSph^{45} \times \mHyp^{5}$, $\mSph^{40} \times \mHyp^{5} \times \mEuc^{5}$, $\mSph^{45} \times \mEuc^{5}$, $\mSph^{50} \times \mEuc^{10}$, $\mSph^{50} \times \mHyp^{10}$, $\mSph^{50} \times \mHyp^{5} \times \mEuc^{5}$, $\mSph^{30} \times \mHyp^{30}$, $\mSph^{30} \times \mEuc^{30}$. We searched learning rates as well as GradNorm~\cite{chen2018gradnorm} weighting parameters within a linear scale. We searched conformality parameter values in the set: $\{0.0, 0.2, 0.4, 0.6, 0.8, 1.0\}$.

We have implemented all algorithms presented in the paper using \texttt{PyTorch}; the code will be made available with scripts to reproduce the results presented here upon acceptance.
Our experiments were run on Amazon AWS instances and were orchestrated using \href{https://spell.run}{Spell}.
Each instance had a single NVidia T4 GPU and 16GB of RAM.\@
Experiments on the \DMammals{} dataset were run for 2,000 epochs and took on average about 16 minutes.
Experiments on the \DNouns{} dataset were run for 1,000 epochs (corresponding to 10,000 steps) and took on average about 5 hours and 45 minutes.
We ran a net total of about 1,000 runs for the results presented in the paper.

\subsection{Proof of Theorem 1}

Recall the constrained and unconstrained loss objectives for conformality defined in Section 3 of the main paper:
\begin{align*}
  \loss_\conformal^{\text{contstr.}}(p) &= 
      \left\| \log\left(F_{p} {G\inS_{p}}\inv e^{-\alpha} \right) \right\|^2 \quad\text{subj.\ to}~-C \le \alpha \le C, \\
  \loss_\conformal^{\text{uncontstr.}}(p)
    &= \left\| \log\left(F_p {G\inS_{p}}\inv \right) \right\|^2 - 
    \begin{cases}
      \frac{1}{n} \log\det(F_p {G\inS_{p}}\inv)^2 & \text{if~} \frac{1}{n} \left|\log\det(F_p {G\inS_{p}}\inv) \right| \le C \\
      2 C \left|\log\det(F_p {G\inS_{p}}\inv) \right| & \text{otherwise}
    \end{cases}.
\end{align*}
Here $F_p$ denotes the pullback metric tensor, $G\inS_{p}$ denotes the source metric tensor, $C$ is the absolute upper bound on the relative difference between these two metric tensors, and $\alpha$ is an additional optimization variable present in the constraints objective.

In the following lemma, we show optimize out $\alpha$ to derive the unconstrained objective.
\begin{lemma}
  For any objective function $J(\theta, \alpha) = \loss(\theta) + \loss_\conformal^\text{constr.}(p)$, $\theta^*$ minimizes $J(\theta, \alpha)$ iff it also minimizes $J'(\theta) = \loss(\theta) + \loss_\conformal^\text{unconstr.}(p)$.
\end{lemma}
\begin{proof}
  In general, we aim to optimize an objective of the form:
  \begin{align*}
    J(\theta, \alpha) &= \loss(\theta) + \left\| \log\left(F_\theta G\inv e^{-\alpha} \right) \right\|^2,
  \end{align*}
  where $|\alpha| < C$, and $\loss(\theta)$ is some function which only depends on $\theta$; for notational clarity, we have dropped $p$ and $\mS$ and highlighted the $\theta$ dependence of $F$.

We can simplify this objective as follows:
\begin{align*}
  J(\theta, \alpha) 
    &= \loss(\theta) + \left\| \log\left(F_\theta G\inv  e^{-\alpha} \right) \right\|^2 \\
    &= \loss(\theta) + \left\| \log\left(F_\theta G\inv \right) - \alpha I  \right\|^2 \\
    &= \loss(\theta) + \left\| \log\left(F_\theta G\inv \right) \right\|^2 - 2 \alpha \tr{\log\left(F_\theta G\inv \right)} + \left\| \alpha I \right\|^2 \\
    &= \loss(\theta) + \left\| \log\left(F_\theta G\inv \right) \right\|^2 - 2 \alpha \log\det\left(F_\theta G\inv \right) + \alpha^2 n,
\end{align*}
where we have used Jacobi's formula: $\tr\log\left(F_\theta G\inv \right) = \log\det\left(F_\theta G\inv \right)$.

For any $\theta$, and an unconstrained $\alpha$, we can solve for the optimal value of $\alpha$:
\begin{align*}
  \frac{\partial J(\theta, \alpha^*)}{\partial \alpha} &= - 2 \log\det\left(F_\theta G\inv \right) + 2 \alpha^* n = 0 \\
  \alpha^* &= \frac{1}{n}\log\det\left(F_\theta G\inv \right).
\end{align*}

Substituting in the original objective, we get:
\begin{align*}
  J(\theta) 
    &= \loss(\theta) + \left\| \log\left(F_\theta G\inv \right) \right\|^2 -
    2 \frac{1}{n} \log\det\left(F_\theta G\inv \right) \log\det\left(F_\theta G\inv \right) + \frac{1}{n^2} \log\det\left(F_\theta G\inv \right)^2 n \\
    &= \loss(\theta) + \left\| \log\left(F_\theta G\inv \right) \right\|^2 - \frac{1}{n} \log\det\left(F_\theta G\inv \right)^2.
\end{align*}

Note that $J(\theta, \alpha)$ is quadratic in $\alpha$: when $\frac{1}{n} \left| \log\det\left(F_\theta G\inv \right) \right| > C$, the optimal value of $J(\theta, \alpha)$ will be at the boundary: $\alpha^* = \operatorname{sgn}\left(\log\det\left(F_\theta G\inv \right)\right) C$. As a result:
\begin{align*}
  J(\theta) 
    &= \loss(\theta) + \left\| \log\left(F_\theta G\inv \right) \right\|^2 - 2 C \left|\log\det\left(F_\theta G\inv \right)\right| + C^2 n.
\end{align*}

Combining these two cases, we get the final result:
\begin{align*}
  J(\theta) 
    &= \loss(\theta) + \left\| \log\left(F_\theta G\inv \right) \right\|^2 - 
    \begin{cases}
      \frac{1}{n} \log\det(F_\theta G\inv)^2 & \text{if~} \left |\frac{1}{n} \log\det(F_\theta G\inv) \right| \le C \\
      2 C \left|\log\det(F_\theta G\inv) \right| & \text{otherwise}
    \end{cases}.
\end{align*}

\end{proof}

\begin{theorem}
     For all values of $C$, $f$ is a conformal map iff $\loss_\conformal(p) = 0$ for all points $p \in \mS$;
    if $C = 0$, then $f$ is a isometric map iff $\loss_\conformal(p) = 0$ for all points $p \in \mS$.
\end{theorem}
\begin{proof}
Suppose $C = 0$. If $\loss_{\conformal}(p) = 0$ for all points $p$ then, $$\left\| \log\left(F_{p} {G\inS_{p}}\inv \right)\right\|^2 = 0$$ for all points $p$. This is true if and only if the geodesic distance between $F_p$ and ${G\inS_{p}}$ on the manifold of positive definite matrices equals zero which is true if and only if $F_p = {G\inS_{p}}$. Having a pullback metric that is equal to the metric on the original space is the definition of an isometric map between Riemannian manifolds. This can be seen to align with the intuitive definition of an isometry---a function that preserves distances. For any two points $p_1, p_2$ we see that the distances are preserved:
\begin{align*}
d\inS(p_1, p_2) &= \min_\gamma \int_{0}^{1} \sqrt{G\inS_{\gamma(t)}(\dot{\gamma}(t), \dot{\gamma}(t))} dt 
\\ &= \min_\gamma \int_{0}^{1} \sqrt{F_{\gamma(t)}(\dot{\gamma}(t), \dot{\gamma}(t))} dt \\
&= \min_\gamma \int_{0}^{1} \sqrt{G\inT_{f(\gamma(t))}(df (\dot{\gamma}(t)), df(\dot{\gamma}(t)))} dt = d \inT (f(p_1), f(p_2))
\end{align*}
where $G \inT$ is the target manifold metric tensor and $\gamma$ is any smooth curve connecting $p_1, p_2$. The Myers–Steenrod theorem gives the converse: that every distance preserving map between Riemannian manifolds is necessarily a smooth isometry of Riemannian manifolds and has a pullback metric equal to the metric on the original space.

Now, suppose $C$ is any value. By similar logic, the loss is nonzero if and only if the geodesic distance between $F_p$ and $e^{\alpha} G\inS_{p}$ is zero. This is true if and only if $F_p = e^{\alpha} G\inS_{p}$ which is the requirement by definition for a map between Riemannian manifolds to be considered conformal. We can see this aligns with the traditional definition of conformal: angles between tangent vectors are preserved as for any tangent vectors $v_1, v_2$ at a point $p$:
\begin{align*}
\cos_G\left(\theta(v_1, v_2)\right) 
&= \frac{G\inS_p(v_1, v_2)}{\sqrt{G\inS_p(v_1, v_1) G\inS_p(v_2, v_2)}} \\
&= \frac{e^{\alpha} G\inS_p(v_1, v_2)}{\sqrt{e^{\alpha} G\inS_p(v_1, v_1) e^{\alpha}G\inS_p(v_2, v_2)}} \\
&= \frac{F_p(v_1, v_2)}{\sqrt{F_p(v_1, v_1) F_p(v_2, v_2)}}
    = \cos_F\left(\theta(v_1, v_2)\right),
\end{align*}
where $\cos_G\left(\theta(v_1, v_2)\right)$ and $\cos_F\left(\theta(v_1, v_2)\right)$ measure the cosine of the angles between $v_1$ and $v_2$ according to the source metric $G\inS_p$ and pullback metric $F_p$.
In other words, any map which preserves angles between any tangent vectors at every point must have a pullback metric that is a positive scalar multiple of the source metric at every point.
\end{proof}

\end{document}